\documentclass[hidelinks,onefignum,onetabnum]{siamart251216}

\usepackage[numbers,sort&compress]{natbib}
\usepackage{amsmath,amssymb,amsfonts,mathtools,bm,bbm}
\usepackage{graphicx,float,algorithm,algpseudocode,placeins}
\usepackage{longtable,wrapfig,rotating}
\usepackage[normalem]{ulem}
\usepackage{capt-of,fancyvrb,multirow}
\usepackage{physics}
\usepackage{tikz,tikz-cd}
\usepackage[textwidth=20mm,textsize=tiny,color=green!40]{todonotes}
\setuptodonotes{inline}
\newcommand{\qf}{\operatorname{qf}}
\newcommand{\MFRes}{\operatorname{MFRes}}
\newsiamthm{dfn}{Definition}
\newsiamthm{assumption}{Assumption}
\newsiamthm{prop}{Proposition}
\newcounter{prop}[section]
\renewcommand{\theprop}{\thesection.\arabic{prop}}
\newcounter{cor}[section]
\renewcommand{\thecor}{\thesection.\arabic{cor}}
\newsiamremark{remark}{Remark}

\headers{Stochastic Dimension Zeroth-Order Estimator}{Z. Liang and H. Gao}
\title{Stochastic Dimension Zeroth-Order Estimator: Memory-Efficient and Stable Training for High-Dimensional and High-Order PINNs%
\thanks{\funding{This work was supported by the National Natural Science Foundation of China under grant 12572138.}}}

\author{Zhangyong Liang\thanks{National Center for Applied Mathematics, Tianjin University, Tianjin 300072, China.}
\and Huanhuan Gao\thanks{School of Mechanical and Aerospace Engineering, Jilin University, Changchun 130025, China (\email{gao\_huanhuan@jlu.edu.cn}).}}

\mathtoolsset{showonlyrefs}

\ifpdf
\hypersetup{
  pdftitle={Stochastic Dimension Zeroth-Order Estimator},
  pdfauthor={Zhangyong Liang and Huanhuan Gao}}
\fi

\begin{document}
\maketitle

\begin{abstract}
The spatial differentiation operations in physics-informed neural networks (PINNs) to solve high-dimensional and high-order partial differential equations (PDEs) always incur heavy memory consumption, which has become the primary computational bottleneck.
Although stochastic dimensional estimators can relieve the memory burden to some extent through randomized differential-operator sampling, the reverse-mode automatic differentiation (AD) still introduces massive memory growth.
The AD-free zeroth-order optimization seems to be the ideal solution to the severe memory occupation problem; however, it will cause gradient variance explosion and further result in unstable training.
To address this issue, we propose the light-weight and stable \textbf{S}tochastic \textbf{D}imension \textbf{Z}eroth-Order \textbf{E}stimator (\textbf{SDZE}), a backpropagation-free framework that couples common-random-number synchronization (CRNS) with layer-wise low-rank perturbations.
CRNS reuses the complete spatial random states and removes the independent-sampling variance singularity term $\mathcal{O}(\varepsilon^{-2})$, while retaining finite-time convergence bounds through stochastic residual derivation.
The matrix-free perturbation can avoid the explicit global subspace basis, the dense perturbation matrices, and the reverse-mode parameter-gradient storage, allowing each parameter update to be computed entirely via paired forward residual evaluations.
In the theoretical analysis, the bounds of pseudo-gradient bias and variance, the layer-wise subspace coverage and the finite-time projected-stationarity guarantees are derived, respectively.
Then the finite-time bounds are yielded under explicit coverage and refresh conditions.
Experiments on high-dimensional and high-order PDEs demonstrate that SDZE can achieve stable training, substantial memory savings, and accuracy comparable to stochastic first-order baselines.
As such, SDZE offers a theoretically grounded, experimentally verified approach that combines randomized spatial-operator estimation with memory-efficient and stable zeroth-order PINN training.
\end{abstract}



\section{Introduction}
\label{sec:intro}

High-dimensional and high-order PDEs often incur rapidly increasing computational and memory costs as the spatial dimension and differential order grow.
Hamilton--Jacobi--Bellman, Fokker--Planck, and Black--Scholes equations are representative high-dimensional problems, whereas Korteweg--de Vries and Cahn--Hilliard equations exemplify PDEs involving high-order spatial derivatives.
Physics-informed neural networks (PINNs) \citep{raissi2019pinn} offer a mesh-free approximation framework for PDEs and complex geometries.
Their scalability is nevertheless constrained by the cost of evaluating high-dimensional and high-order spatial operators, as well as the reverse-mode automatic differentiation (AD) for parameter updates.
High-order forward-mode AD extends the forward sweep for spatial-derivative evaluation \citep{bendtsen1997tad,karczmarczuk1998ad}.
General frameworks further support higher-order differentiation \citep{wang2017higherad,laurel2022higherad}.
Taylor-mode AD has been incorporated into machine-learning workflows \citep{bettencourt2019taylor}.
Operator-specific forward rules reduce the cost of Laplacian evaluation \citep{li2024computational} and related differential operators \citep{li24_dof}.
Randomized linearizations offer another way to reduce the cost of an AD graph \citep{oktay2021randomad}.
Coupled automatic--numerical differentiation likewise accelerates PINN residual evaluation while retaining derivative information \citep{chiu2022can}.

\begin{figure}[htbp]
  \centering
  \includegraphics[width=\linewidth]{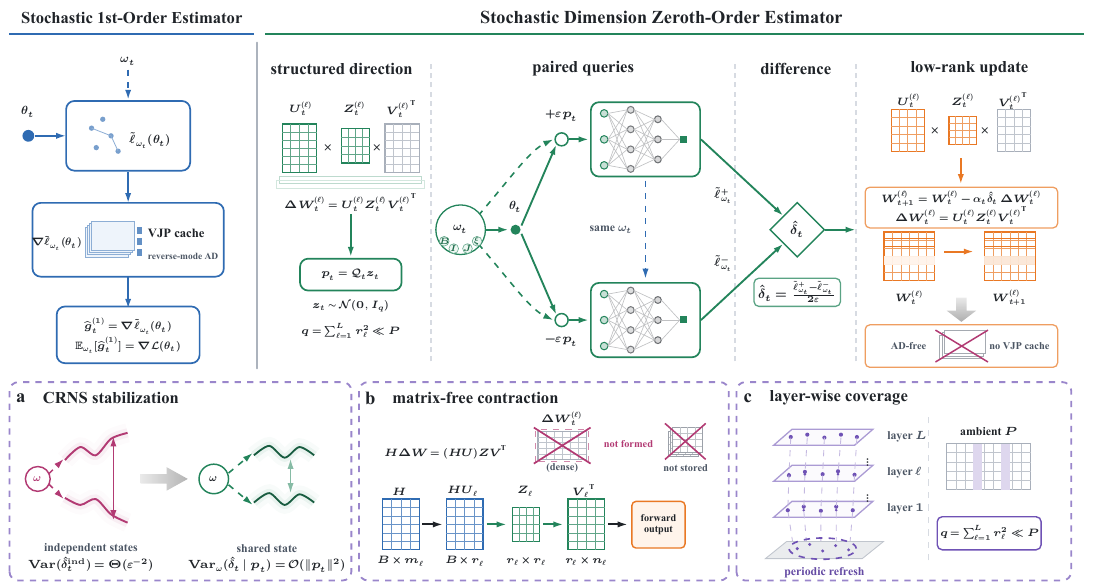}
  \vspace{-12pt}
  \caption{Framework of SDZE.
  The top row contrasts stochastic first-order and AD-free updates.
  The bottom row shows \textbf{a}, CRNS variance control; \textbf{b}, matrix-free low-rank contraction; and \textbf{c}, layer-wise coverage with periodic refresh.}
\label{fig:sdze-framework}
\end{figure}

To scale up PINNs for high-dimensional and high-order PDEs with a suitable operator structure, dimension-independent stochastic spatial estimators have been developed to amortize the computational cost.
Randomized numerical linear algebra motivates operator estimators that avoid enumerating all high-dimensional terms \citep{martinsson2020randomized,murray2023rnla}.
Random projections provide a complementary route to dimension reduction \citep{ghojogh2021jl}.
For example, Stochastic Dimension Gradient Descent (SDGD) \citep{hu2024sdgd} mitigates the high-dimensional bottleneck by randomizing over input dimensions.
Score-PINN takes the advantage of the randomized score-matching constructions for high-dimensional Fokker--Planck equations \citep{hu2025score}, whereas Hutchinson Trace Estimation (HTE) introduces random Rademacher or Gaussian masks to transform large Hessian or Jacobian evaluations into stochastic vector products \citep{hutchinson1989trace,hu2024hutch}.
The Stochastic Taylor Derivative Estimator (STDE) \citep{shi2024stochastic,shi2026stde++} applies forward Taylor-mode AD with sparse random jets to contract high-order differential operators.
Stochastic sampling has also been used for fractional and tempered fractional operators \citep{hu2024tackling,guo2022monte}.
Separable neural operators reduce the effective dimension of physics-informed models through structured representations \citep{mandl2025separable}.
Overall, these methods can help reduce the cost of estimating high-dimensional and high-order spatial operators; however, they still rely on backpropagation to update network parameters.
Despite the backpropagation process, stochastic spatial estimators still inherit the large memory cost of first-order (FO) parameter optimization \citep{amari1993backpropagation,kingma15_adam}.

At the parameter-update level, two-point ZO methods estimate gradients from paired forward evaluations \citep{ghadimi2013stochastic,liu2020primer}.
Zeroth-order optimization removes backpropagation from parameter updates.
In large-language-model fine-tuning, ZO can reduce the incremental memory requirement to that of forward evaluation \citep{malladi2023fine}.
Full-space ZO estimators, however, include the variance, which increases with the number of trainable parameters and brings forward the subspace perturbations in large models \citep{gautamvariance, liu2026sparse}.
Finite-difference accuracy balances bias and variance \citep{nesterov2017random,duchi2015optimal}.
The ZO variance can be reduced by increasing the training batch size or restricting the perturbation to fewer parameter directions \citep{yue2023zeroth,jiang2024zo}.
Typical measures such as sparse masks \citep{liu2026sparse} and parameter-efficient tuning \citep{zhang2024revisiting,malladi2023fine} implement the above strategies.
Tensorized adapters \citep{yang2024adazeta} provide another implementation.
Alternatively, stochastic low-dimensional subspaces offer other options \citep{nozawa2024zeroth,roberts2023direct,kozak2021stochastic}, whose explicit basis will lead to $\mathcal{O}(P\times r)$ storage, and the storage problem may result in the impracticality for extremely high-dimensional PINNs.
Moreover, directly combining ZO finite differences with stochastic spatial operators will introduce a divergent source of the estimation variance, in which two independently sampled spatial residuals are subtracted and divided by a small perturbation step $\epsilon$, producing an $\mathcal{O}(\epsilon^{-2})$ term.
Common random numbers couple paired stochastic evaluations for variance reduction \citep{glasserman2003monte,robinson1996analysis}.

To address this issue, we firstly put forward the \textbf{S}tochastic \textbf{D}imension \textbf{Z}eroth-\textbf{O}rder \textbf{E}stimator (\textbf{SDZE}), a backpropagation-free training strategy that uses the common-random-number synchronization (CRNS) and layer-wise low-rank perturbations for stochastic spatial PINN solvers with forward-evaluable residual losses \citep{yu2025zeroth}.
The proposed SDZE avoids storing reverse-mode parameter-gradient traces and admits finite-time projected-stationarity bounds governed by the subspace dimension $q$.
The advantage of SDZE is even more outstanding, especially when the out-of-memory (OOM) issue makes FO updates impractical.
Figure~\ref{fig:sdze-framework} presents the stochastic FO and SDZE update schemes.

The main contributions of this work are summarized as follows:
\begin{itemize}
    \item \textbf{CRNS for stochastic ZO estimators.} We discover the $\mathcal{O}(\epsilon^{-2})$ variance term that leads to unstable training when independently sampled stochastic spatial residuals are used in two-sided ZO finite differences.
        CRNS reuses the complete random state across the paired evaluations and eliminates this independent-sampling variance singularity, guaranteeing the training stability.
        \item \textbf{Matrix-free layer-wise subspace perturbations.} We derive an implicit forward-pass formulation through the related tensor contraction.
        The formulation retains the model parameters while avoiding the $P\times q$ subspace basis matrix, the $m_l\times n_l$ perturbation matrix, and the reverse-mode parameter-gradient storage.
        \item \textbf{Variance and convergence analysis.} Under stated regularity conditions, we bound the coupled spatial finite-difference variance under CRNS and establish finite-time projected-stationarity guarantees.
        Building on this analysis, we further characterize the effects of finite-difference bias, stochastic residual noise, layer-wise subspace coverage, and derive full-space convergence bounds under explicit coverage and refresh conditions.
        \item \textbf{Backpropagation-free limited-memory updates.} SDZE executes the parameter-update step with stochastic spatial solvers such as the SDGD, the HTE, and the STDE when the corresponding residual losses are forward-evaluable.
        This strategy can help remove the reverse-mode AD and light-weight the GPU memory burden of the first-order updates.
        \item \textbf{Memory-efficient and variance-stable training.} We evaluate SDZE on high-dimensional and high-order PDE benchmarks.
        The results demonstrate that CRNS suppresses variance amplification in stochastic ZO updates, while SDZE reduces peak GPU memory and achieves accuracy comparable to executable stochastic FO spatial estimators.
\end{itemize}

The paper is organized as follows.
Section~\ref{sec:prelim} formulates the PINN optimization problems.
Section~\ref{sec:method} defines the stochastic residuals, the coupled ZO estimator CRNS, and the matrix-free strategy.
Section~\ref{sec:theory} gives the analytical results of the ZO estimator.
Numerical experiments are presented in Section~\ref{sec:exp}, followed by the conclusions in Section~\ref{sec:conclusion}.

\section{Preliminaries} 
\label{sec:prelim}

\subsection{Notation}
Scalars, vectors, and matrices are denoted by non-bold, bold lowercase, and bold uppercase letters, respectively.
We use $\mathrm{vec}(\boldsymbol W)$ for column-wise vectorization and $\otimes$ for the Kronecker product.
The standard Gaussian distribution is $\mathcal N(\boldsymbol0,\boldsymbol I)$, and $\mathbb E$ and $\mathrm{Var}$ denote expectation and variance.
The notation $\|\cdot\|$ denotes the Euclidean or spectral norm as appropriate, and $\|\cdot\|_F$ denotes the Frobenius norm.
Let $\boldsymbol\theta\in\mathbb R^P$ denote all trainable parameters.

\subsection{First-order gradients}
\label{sec:first-order-gradient-estimation}
First-order automatic differentiation (AD) evaluates derivatives through JVPs or VJPs.
For a depth-$L$, width-$h$ network, a full $d$-dimensional input Jacobian requires $d$ JVPs, whereas reverse-mode AD stores the forward trace and requires $\mathcal{O}(d+(L-1)h)$ memory.
Repeated AD for $k$th-order input derivatives therefore retains an exponential dependence on $k$.

\subsection{Zeroth-order gradient estimation}
\label{sec:zo-gradient}

Zeroth-order (ZO) estimation replaces parameter VJPs with paired loss evaluations.
For a minibatch $\mathcal{B}$, the two-point Gaussian estimator is
\[
\widehat{\nabla} \mathcal{L}(\boldsymbol{\theta}; \mathcal{B}) = \frac{\mathcal{L}(\boldsymbol{\theta} + \epsilon \boldsymbol{z}; \mathcal{B}) - \mathcal{L}(\boldsymbol{\theta} - \epsilon \boldsymbol{z}; \mathcal{B})}{2 \epsilon} \boldsymbol{z},
\]
where $\boldsymbol{z}\sim\mathcal{N}(\boldsymbol{0},\boldsymbol I_P)$ and $\epsilon>0$.
Full-space perturbations have variance scaling with $P$.
Using an orthonormal $\mathcal Q\in\mathbb R^{P\times q}$ with $q\ll P$ gives
\[
\widehat{\nabla} \mathcal{L}(\boldsymbol{\theta}, \mathcal{Q}; \mathcal{B}) = \frac{\mathcal{L}(\boldsymbol{\theta} + \epsilon \mathcal{Q} \boldsymbol{z}; \mathcal{B}) - \mathcal{L}(\boldsymbol{\theta} - \epsilon \mathcal{Q} \boldsymbol{z}; \mathcal{B})}{2 \epsilon} \mathcal{Q} \boldsymbol{z}.
\]
Its sampling variance depends on $q$, but explicitly storing $\mathcal Q$ costs $\mathcal O(Pq)$ memory.

\subsection{Stochastic dimension gradient descent}
\label{sec:sdgd-review}
SDGD \cite{hu2024sdgd} samples terms of high-dimensional additive differential operators.
For $\mathcal{D} = \sum_{j=1}^{N_{\mathcal{D}}} \mathcal{D}_j$, it uses
\[
\tilde{\mathcal{D}}_J = \frac{N_{\mathcal{D}}}{|J|} \sum_{j \in J} \mathcal{D}_j,
\]
where $J$ is a uniform subset of operator terms.
Treating non-sampled dimensions as constants reduces the AD memory from $\mathcal{O}(2^{k-1}(d+(L-1)h))$ to $\mathcal{O}(|J|2^{k-1}(1+(L-1)h))$, while the exponential dependence on derivative order $k$ remains.

\section{Method}
\label{sec:method}

To address the $\mathcal{O}(d^k)$ cost of expanded high-order spatial operators and the $\mathcal{O}(P)$ memory cost of reverse-mode automatic differentiation (AD), we introduce SDZE, which combines randomized spatial operators with matrix-free low-rank subspace zeroth-order updates.
CRNS shares the stochastic spatial state between finite-difference evaluations, removing the singular $\mathcal{O}(\epsilon^{-2})$ term from independent residuals while retaining finite gradient noise.

\begin{figure}[t]
  \centering
  \includegraphics[width=\linewidth]{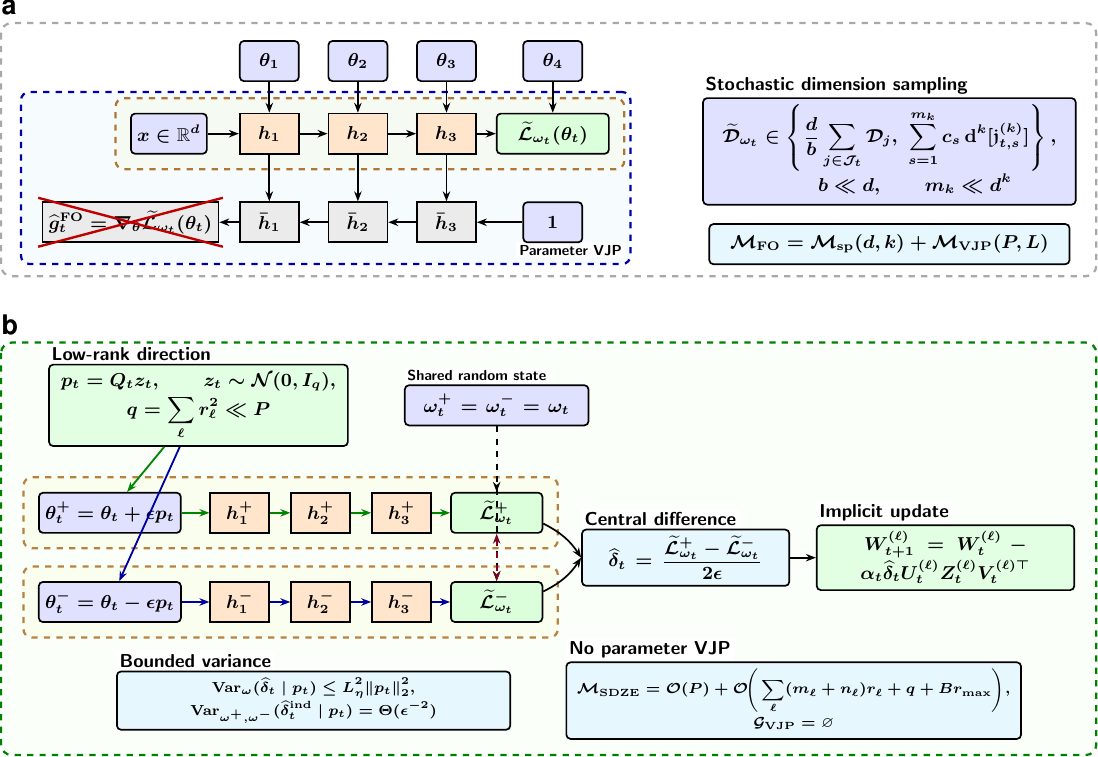}
  \vspace{-16pt}
  \caption{First-order and SDZE updates.
  \textbf{a}, first-order updates propagate parameter VJPs.
  \textbf{b}, SDZE evaluates paired finite-difference losses with a shared CRNS state.}
  \label{fig:sdze-update-paths}
\end{figure}

The stochastic first-order and SDZE parameter-update paths are compared in Figure~\ref{fig:sdze-update-paths}.
The first-order update includes a parameter vector--Jacobian-product graph.
Under SDZE, the two loss evaluations share a common random state, and the low-rank update is obtained from the finite-difference loss difference and layer-wise factors.
The layer-wise subspace construction follows stochastic zeroth-order estimation \cite{yu2025zeroth}.
The two loss evaluations are constructed with CRNS according to a common-random-number variance-reduction principle \cite{glasserman2003monte,spall2003introduction,robinson1996analysis}.
Related stochastic-approximation analyses are given in \cite{kushner2003stochastic,berahas2022theoretical}.

Algorithm~\ref{alg:sdze} provides an overview of the SDZE training workflow.
At each iteration, SDZE samples layer-wise low-rank factors, evaluates paired residuals with a shared state, and applies a matrix-free update.
Here, \(\MFRes(\cdot)\) denotes the stochastic residual at the signed perturbation, and \(\qf(\boldsymbol R)\) the sign-normalized \(Q\) factor from the QR decomposition of \(\boldsymbol R\).
Homogeneous augmentation incorporates biases into \(\boldsymbol W^{(l)}\), whose row dimension is \(m_l\).


\begin{algorithm}[t]
  \caption{SDZE training process}
  \label{alg:sdze}
  \begin{algorithmic}[1]
    \Require \(\boldsymbol\theta_0=\{\boldsymbol W_0^{(l)}\}_{l=1}^{L}\), ranks \(\{r_l\}_{l=1}^{L}\), refresh period \(F\), schedules \(\{\epsilon_t,\alpha_t\}_{t=0}^{T-1}\), and row partitions \(\{\mathcal S^{(l)}\}_{l=1}^{L}\).
    \Ensure Parameters \(\boldsymbol\theta_T\)
    \For{\(t=0,\ldots,T-1\)}
      \If{\(t \bmod F=0\)}
        \For{\(l=1,\ldots,L\)}
          \State Sample independent \(\boldsymbol R_{U,t}^{(l)}\) and \(\boldsymbol R_{V,t}^{(l)}\) with i.i.d. Gaussian entries
          \State \(\boldsymbol U_t^{(l)}\gets\qf(\boldsymbol R_{U,t}^{(l)})\), \(\boldsymbol V_t^{(l)}\gets\qf(\boldsymbol R_{V,t}^{(l)})\)
        \EndFor
      \Else
        \State Reuse \((\boldsymbol U_t^{(l)},\boldsymbol V_t^{(l)})\gets(\boldsymbol U_{t-1}^{(l)},\boldsymbol V_{t-1}^{(l)})\) for all \(l\)
      \EndIf
      \For{\(l=1,\ldots,L\)}
        \State Sample \(\operatorname{vec}(\boldsymbol Z_t^{(l)})\sim\mathcal N(\boldsymbol 0,\boldsymbol I_{r_l^2})\)
      \EndFor
      \State Draw one complete stochastic state \(\omega_t\sim\mathbb P_s\)
      \State \(\widetilde\ell_t^{+}\gets\MFRes(\boldsymbol\theta_t,+\epsilon_t;\omega_t,\mathcal U_t,\mathcal Z_t,\mathcal V_t)\)
      \State \(\widetilde\ell_t^{-}\gets\MFRes(\boldsymbol\theta_t,-\epsilon_t;\omega_t,\mathcal U_t,\mathcal Z_t,\mathcal V_t)\) using the same \(\omega_t\)
      \State \(\widehat\delta_t\gets(\widetilde\ell_t^{+}-\widetilde\ell_t^{-})/(2\epsilon_t)\)
      \For{\(l=1,\ldots,L\)}
        \State \(\boldsymbol C_t^{(l)}\gets\boldsymbol Z_t^{(l)}{\boldsymbol V_t^{(l)}}^\top\)
        \For{\(S_k\in\mathcal S^{(l)}\)}
          \State \(\boldsymbol W_{t+1}^{(l)}[S_k,:]\gets\boldsymbol W_t^{(l)}[S_k,:]-\alpha_t\widehat\delta_t\boldsymbol U_t^{(l)}[S_k,:]\boldsymbol C_t^{(l)}\)
        \EndFor
      \EndFor
    \EndFor
    \State \Return \(\boldsymbol\theta_T\)
\end{algorithmic}
\end{algorithm}

\subsection{Stochastic operator and residual estimators}
\label{sec:method-spatial}

Consider a PDE on a physical domain $\mathcal{X}\subset\mathbb{R}^d$ with the form
\begin{equation}
\mathcal{N}[u]=\sum_{i=1}^{N_{\mathrm{op}}}\mathcal{D}_i[u]+\mathcal{R}[u]=f,
\end{equation}
where the additively decomposable part $\sum_i\mathcal{D}_i$ is randomized and $\mathcal{R}$ is evaluated exactly.
Let $u_{\boldsymbol{\theta}}:\mathbb{R}^d\to\mathbb{R}^{d'}$ be a PINN with parameters $\boldsymbol{\theta}\in\mathbb{R}^P$, and we define the pointwise residual objective and the corresponding statistical objective by
\begin{equation}
\ell(\boldsymbol{\theta};\boldsymbol{x})=
\frac12\left\|\mathcal{N}[u_{\boldsymbol{\theta}}](\boldsymbol{x})-f(\boldsymbol{x})\right\|_2^2,
\qquad
\mathcal{L}(\boldsymbol{\theta})=\mathbb{E}_{\boldsymbol{x}\sim\mu_{\mathcal X}}
\left[\ell(\boldsymbol{\theta};\boldsymbol{x})\right].
\label{eq:pointwise-objective}
\end{equation}

Note that the formulation in \eqref{eq:pointwise-objective} can be applied only when the randomized component has the stated additive structure.
The derivatives in $\mathcal{D}_i$ and $\mathcal{R}$ are evaluated by the spatial-derivative evaluation method.
Taylor-mode AD deterministically evaluates derivatives, whereas STDE provides randomized spatial-operator estimation.

Let $(\Xi,\mathcal{F},\mathbb{P}_s)$ be the probability space governing the random state of the randomized spatial-operator estimator, and let $\omega\in\Xi$ denote any one complete random state.
For an additive operator, a state sampled uniformly with $b=|\mathcal I|$ may contain an index subset $\mathcal I\subset\{1,\ldots,N_{\mathrm{op}}\}$.
The randomized spatial-operator estimator is expressed as
\begin{equation}
\widetilde{\mathcal N}_{\mathcal I}[u]
=\frac{N_{\mathrm{op}}}{b}\sum_{i\in\mathcal I}\mathcal D_i[u]+\mathcal R[u],
\qquad
\mathbb{E}_{\mathcal I}[\widetilde{\mathcal N}_{\mathcal I}[u]]=\mathcal N[u].
\label{eq:spatial_oracle}
\end{equation}

For two independent spatial states $\omega_1,\omega_2\overset{\mathrm{i.i.d.}}{\sim}\mathbb P_s$, write $\omega=(\omega_1,\omega_2)$.
The cross-sampled quantity
\begin{equation} \label{eq:stoch_loss}
\widetilde\ell_{\omega}(\boldsymbol{\theta};\boldsymbol{x})
=\frac12\widetilde{\boldsymbol r}_{\omega_1}(\boldsymbol\theta;\boldsymbol x)^\top
\widetilde{\boldsymbol r}_{\omega_2}(\boldsymbol\theta;\boldsymbol x),
\qquad
\widetilde{\boldsymbol r}_{\omega_j}
=\widetilde{\mathcal N}_{\omega_j}[u_{\boldsymbol\theta}](\boldsymbol x)-f(\boldsymbol x),
\end{equation}
is an unbiased stochastic estimator of $\ell(\boldsymbol\theta;\boldsymbol x)$ in which individual realizations may take either sign.
From the parameter-optimization viewpoint, the mapping $(\boldsymbol\theta,\boldsymbol x,\omega)\mapsto\widetilde\ell_\omega(\boldsymbol\theta;\boldsymbol x)$ defines a stochastic zeroth-order residual oracle.
Each query returns the scalar residual objective with the specified parameters, collocation point, and random state.
Define $\eta_\omega(\boldsymbol\theta;\boldsymbol x)=\widetilde\ell_\omega(\boldsymbol\theta;\boldsymbol x)-\ell(\boldsymbol\theta;\boldsymbol x)$, then $\mathbb{E}_\omega[\eta_\omega]=0$.
We assume that the corresponding variances of the $\eta_\omega$ are finite.
The lower bounds depend on the spatial-noise level and the sampling resources.

\subsection{Layer-wise subspaces and refresh}
\label{sec:method-subspace}

SDZE uses dynamic layer-wise low-rank subspaces to avoid parameter-gradient storage.
The variance of a full-space isotropic direction grows with the ambient parameter dimension.
The active subspace has dimension $q=\sum_l r_l^2$.
For the $l$-th parameter matrix $\boldsymbol{W}^{(l)} \in \mathbb{R}^{m_l \times n_l}$, we choose the rank $r_l\leq\min(m_l,n_l)$.
Then the shape factors are $\boldsymbol U_t^{(l)}\in\mathbb R^{m_l\times r_l}$ and $\boldsymbol V_t^{(l)}\in\mathbb R^{n_l\times r_l}$.
The implementation preserves the matrix layout and remains matrix-free.

The subspace diversity and QR cost are balanced by the \emph{Periodic Lazy Subspace Refresh} with a fixed frequency $F\ge1$.
At an update step, the independent standard Gaussian matrices are sampled for both factors of each layer.
Then the QR decomposition with a sign-normalized diagonal of the $R$ factor yields the basis with the standard Haar distribution on the relevant Stiefel manifolds.
Refreshing the two factors of layer $l$ costs $\mathcal O((m_l+n_l)r_l^2)$ operations.
Amortized over one refresh block, this memory cost is $\mathcal O((m_l+n_l)r_l^2/F)$ per iteration.
As a result, each updated basis is independent of the history before the refresh block.
At training step $t$,
\begin{equation}
\left( \boldsymbol{U}^{(l)}_t, \boldsymbol{V}^{(l)}_t \right) = 
\begin{cases}
    \left(\qf(\boldsymbol{R}_{1,t}^{(l)}),\qf(\boldsymbol{R}_{2,t}^{(l)})\right) & \text{if } t \equiv 0 \pmod F, \\
    \left( \boldsymbol{U}^{(l)}_{t-1}, \boldsymbol{V}^{(l)}_{t-1} \right) & \text{otherwise},
\end{cases}
\label{eq:refresh}
\end{equation}
In \eqref{eq:refresh}, $\boldsymbol{R}_{1,t}^{(l)}$ and $\boldsymbol{R}_{2,t}^{(l)}$ are standard Gaussian matrices.
The QR factors satisfy
\[
  {\boldsymbol{U}^{(l)}_t}^\top \boldsymbol{U}^{(l)}_t=\boldsymbol{I}_{r_l},
  \qquad
  {\boldsymbol{V}^{(l)}_t}^\top \boldsymbol{V}^{(l)}_t=\boldsymbol{I}_{r_l}.
\]

At each step, the low-dimensional core perturbation matrix $\boldsymbol{Z}^{(l)}_t \in \mathbb{R}^{r_l \times r_l}$ is independently sampled with $\mathrm{vec}(\boldsymbol{Z}^{(l)}_t) \sim \mathcal{N}(\boldsymbol{0}, \boldsymbol{I}_{r_l^2})$.
The random perturbation direction within the active subspace can be refreshed by the sampling at each step.

The layer-wise and global perturbations are
\begin{equation} \label{eq:kronecker}
\begin{aligned}
    \Delta \boldsymbol{W}^{(l)}_t=\boldsymbol{U}^{(l)}_t\boldsymbol{Z}^{(l)}_t{\boldsymbol{V}^{(l)}_t}^\top, \qquad
    \Delta \boldsymbol{\theta}_t = \mathcal{Q}_t \boldsymbol{z}_t,
\end{aligned}
\end{equation}
where
\begin{equation*}
\left\{
\begin{aligned}
\mathcal{Q}_t
&=\mathrm{diag}\!\left(
  \boldsymbol{V}^{(1)}_t\otimes\boldsymbol{U}^{(1)}_t,
  \dots,
  \boldsymbol{V}^{(L)}_t\otimes\boldsymbol{U}^{(L)}_t\right)
  \in\mathbb{R}^{P\times q},\\
\boldsymbol{z}_t
&=[\mathrm{vec}(\boldsymbol{Z}_t^{(1)})^\top,
  \dots,
  \mathrm{vec}(\boldsymbol{Z}_t^{(L)})^\top]^\top
  \sim\mathcal{N}(\boldsymbol{0},\boldsymbol{I}_q).
\end{aligned}
\right.
\end{equation*}

The global basis $\mathcal{Q}_t$ is introduced only for analysis; the implementation operates directly with the layer-wise factors.
The global direction in \eqref{eq:kronecker} has active dimension $q = \sum_{l=1}^L r_l^2$, and the Kronecker mixed-product property gives $\mathcal{Q}_t^\top \mathcal{Q}_t = \boldsymbol{I}_q$.
Then, $\mathcal{Q}_t\mathcal{Q}_t^\top$ is the corresponding orthogonal projector.

\subsection{CRNS-coupled zeroth-order estimator}
\label{sec:method-crns}
The paired residual queries use \eqref{eq:stoch_loss}.
Independent states in the two-sided perturbation produce a variance term scaling as $\mathcal O(\epsilon^{-2})$.
For a two-sided directional difference evaluated at $(\boldsymbol\theta\pm\epsilon\mathcal Q_t\boldsymbol z_t)$ with independent states $\omega^+$ and $\omega^-$, where $\sigma_\eta^2(\boldsymbol\vartheta):=\operatorname{Var}_{\omega}[\eta_\omega(\boldsymbol\vartheta)]$, the spatial variance is
\begin{equation} \label{eq:independent-sampling-variance}
\operatorname{Var}_{\omega^+,\omega^-}\!\left(
\widehat\delta_t^{\mathrm{ind}}\mid\mathcal Q_t\boldsymbol z_t\right)
=\frac{\sigma_\eta^2(\boldsymbol\theta+\epsilon\mathcal Q_t\boldsymbol z_t)
+\sigma_\eta^2(\boldsymbol\theta-\epsilon\mathcal Q_t\boldsymbol z_t)}{4\epsilon^2}.
\end{equation}

The variance term in \eqref{eq:independent-sampling-variance} diverges as $\epsilon\to0$ when the spatial-noise variance remains nonzero.
Under CRNS, the opposing evaluations use the same complete state $\omega$.
The directional estimator and parameter update are
\begin{equation} \label{eq:delta}
\hat\delta_t=\frac{\widetilde\ell_{\omega}(\boldsymbol\theta+\epsilon\mathcal Q_t\boldsymbol z_t;\boldsymbol x)-\widetilde\ell_{\omega}(\boldsymbol\theta-\epsilon\mathcal Q_t\boldsymbol z_t;\boldsymbol x)}{2\epsilon},
\qquad
\boldsymbol\theta_{t+1}=\boldsymbol\theta_t-\alpha_t\hat\delta_t\mathcal Q_t\boldsymbol z_t.
\end{equation}

Under the mean-square regularity condition, CRNS replaces the inverse-square term in \eqref{eq:independent-sampling-variance} with finite residual-gradient noise.

\subsection{Matrix-free forward evaluation}
\label{sec:method-implicit}

The global basis $\mathcal{Q}_t \in \mathbb{R}^{P \times q}$ is needed for the theoretical abstraction, but its explicit form requires $\mathcal{O}(Pq)$ memory.
At the same time, the explicit layer perturbation $\Delta \boldsymbol{W}^{(l)}$ additionally demands an $\mathcal{O}(m_l n_l)$ memory storage.
For extreme-dimensional PDE inputs (e.g., $m_1 = 10^7$), either representation can exceed the available GPU memory.

The tensor transmission through the layer-wise factors occurs during the evaluation of the shape perturbations.
Let $\boldsymbol H_+^{(l-1)}$ and $\boldsymbol H_-^{(l-1)}$ denote the two sign-specific hidden states, each in $\mathbb R^{B\times m_l}$, and $\sigma(\cdot)$ denote the activation function, the perturbed forward pass at layer $l$ is the recursive contraction
\begin{equation} \label{eq:implicit}
\underbrace{\boldsymbol{H}_{\pm}^{(l)}}_{B \times n_l} = \sigma \Biggl( \underbrace{\boldsymbol{H}_{\pm}^{(l-1)} \boldsymbol{W}^{(l)}}_{B \times n_l} \pm \epsilon \underbrace{\Big( \underbrace{{\color{blue}(\boldsymbol{H}_{\pm}^{(l-1)} \boldsymbol{U}^{(l)}_t)}}_{B \times r_l} \boldsymbol{Z}^{(l)}_t \Big)}_{B \times r_l} {\boldsymbol{V}^{(l)}_t}^\top \Biggr).
\end{equation}

The inner contraction $\boldsymbol H_{\pm}^{(l-1)}\boldsymbol U_t^{(l)}$ has the size $B\times r_l$.
No dense $m_l\times n_l$ perturbation matrix is formed in \eqref{eq:implicit}; the perturbation is evaluated through the associative low-rank contractions.
The memory consumption also includes the ordinary forward-pass activations and the QR factors $\boldsymbol U_t^{(l)}$ and $\boldsymbol V_t^{(l)}$.
Consequently, the dense perturbation matrix and the reverse-mode parameter-gradient information will not be stored.
When the factors are stored rather than reconstructed from seeds, the parameter-side storage consists of the unavoidable $\mathcal O(P)$ model parameters and $\mathcal O\!\left(\sum_l(m_l+n_l)r_l+\sum_l r_l^2\right)$ low-rank factors.

\subsection{Computational realization of SDZE}
\label{sec:method-clarifications}
We define the double-sampled residual and synchronized random state before specifying the paired forward pass, block update, and bias treatment.
\def\SDZEMainTheory{}
%
%

\ifdefined\SDZEMainTheory


\subsubsection{Stochastic residual construction}
For any a collocation point $\boldsymbol{x}$ and let $N:=N_{\mathcal L}$, then we define
\begin{equation}
\boldsymbol a_i(\boldsymbol\theta)
:=\mathcal D_{\boldsymbol x}^{(i)}u_{\boldsymbol\theta}(\boldsymbol x),
\qquad
\boldsymbol A(\boldsymbol\theta):=\sum_{i=1}^{N}\boldsymbol a_i(\boldsymbol\theta),
\qquad
\boldsymbol r(\boldsymbol\theta):=
\boldsymbol A(\boldsymbol\theta)-\boldsymbol f(\boldsymbol x).
\label{eq:det-exact-residual}
\end{equation}
\par
For the exact residual in \eqref{eq:det-exact-residual}, uniform sampling without replacement selects a subset $I\subset\{1,\ldots,N\}$ of size $b$, and we set
\begin{equation}
\widetilde{\boldsymbol A}_{I}(\boldsymbol\theta)
:=\frac{N}{b}\sum_{i\in I}\boldsymbol a_i(\boldsymbol\theta),
\qquad
\boldsymbol r_I(\boldsymbol\theta)
:=\widetilde{\boldsymbol A}_{I}(\boldsymbol\theta)-\boldsymbol f(\boldsymbol x),
\label{eq:det-random-residual}
\end{equation}
where $\mathbb{E}_I[\widetilde{\boldsymbol A}_I]=\boldsymbol A$ and $\mathbb{E}_I[\boldsymbol r_I]=\boldsymbol r$, which proves that the estimator in \eqref{eq:det-random-residual} is unbiased.

Using the same residual realization in both factors produces the positive bias
\begin{equation}
\mathbb{E}_I\!\left[\frac12\|\boldsymbol r_I\|_2^2\right]
=\frac12\|\boldsymbol r\|_2^2
+\frac12\operatorname{tr}
\operatorname{Cov}_I(\widetilde{\boldsymbol A}_I).
\label{eq:det-single-sample-loss-bias}
\end{equation}
\par
Cross-sampling provides an unbiased estimator of the squared PINN residual rather than only an unbiased differential-operator estimator.
Let $I$ and $J$ be independent simple random $b$-subsets, each sampled uniformly without replacement from $\{1,\ldots,N\}$.
The cross-sampled loss is exactly unbiased:
\begin{equation}
\mathbb{E}_{I,J}\!\left[
\frac12\boldsymbol r_I^{\top}\boldsymbol r_J\right]
=\frac12\boldsymbol r^{\top}\boldsymbol r.
\label{eq:det-double-sampling-unbiased}
\end{equation}
\par
The independence between the sampled index sets $I$ and $J$ gives $\mathbb{E}_{I,J}[\boldsymbol r_I^{\top}\boldsymbol r_J] =(\mathbb{E}_I[\boldsymbol r_I])^{\top} (\mathbb{E}_J[\boldsymbol r_J])$.
If the same subset is reused in both factors ($J=I$), the covariance term in \eqref{eq:det-single-sample-loss-bias} remains nonzero.
Random overlap between independently sampled $I$ and $J$ does not affect \eqref{eq:det-double-sampling-unbiased}.
The similar independence argument applies after averaging over a collocation minibatch.

For the statements involving exact gradients, assume the domination condition
\begin{equation}
\sup_{\boldsymbol\vartheta\in\mathcal N}
\|\nabla_{\boldsymbol\vartheta}
\widetilde\ell_{\omega}(\boldsymbol\vartheta)\|_2
\leq G(\omega),
\qquad
\mathbb{E}_{\omega}[G(\omega)]<\infty.
\label{eq:det-dominated-gradient}
\end{equation}
\par
Under \eqref{eq:det-dominated-gradient}, applying dominated convergence to the difference quotient gives
\begin{equation}
\nabla\mathcal L(\boldsymbol\theta)
=\nabla\mathbb{E}_{\omega}[\widetilde\ell_{\omega}(\boldsymbol\theta)]
=\mathbb{E}_{\omega}[\nabla\widetilde\ell_{\omega}(\boldsymbol\theta)].
\label{eq:det-gradient-interchange}
\end{equation}
\par
\eqref{eq:det-gradient-interchange} is an analytical identity, instead of an operational gradient computation.
Then the parameter update can be obtained through the finite-difference evaluation of the residual loss.

Define the finite-population covariance
\begin{equation}
\boldsymbol\Sigma_a
:=\frac{1}{N-1}
\sum_{i=1}^{N}
(\boldsymbol a_i-\overline{\boldsymbol a})
(\boldsymbol a_i-\overline{\boldsymbol a})^{\top},
\qquad
\overline{\boldsymbol a}:=\frac1N\sum_{i=1}^{N}\boldsymbol a_i.
\label{eq:det-pop-cov-def}
\end{equation}
\par
Then \eqref{eq:det-pop-cov-def} yields
\begin{equation}
\operatorname{Cov}_{I}(\widetilde{\boldsymbol A}_{I})
=\frac{N^2}{b}\left(1-\frac{b}{N}\right)
\boldsymbol\Sigma_a.
\label{eq:det-finite-pop-cov}
\end{equation}
\par
The covariance in \eqref{eq:det-finite-pop-cov} retains the factor $1-b/N$ because the sampled fraction $b/N$ cannot be asymptotically neglected.
The derivations of \eqref{eq:det-single-sample-loss-bias} and \eqref{eq:det-finite-pop-cov} are given in the supplementary material.

\begin{remark}
Let $\omega\sim\mathbb P_s$ denote the complete random state of one stochastic residual loss, including the collocation minibatch, independent cross-sampling subsets $(I,J)$, and all randomized spatial probes.
The subsets $(I,J)$ and the two probe copies within each cross-sampled loss remain internally independent.
Under CRNS, the same complete state $\omega$ is used at $\boldsymbol\theta\pm\epsilon\boldsymbol p$; synchronizing only part of the state leaves $\mathcal O(\epsilon^{-2})$ terms.
Uniform subset scaling yields \eqref{eq:spatial_oracle} for SDGD, while HTE and STDE are unbiased under $\mathbb{E}[\boldsymbol v\boldsymbol v^{\top}]=\boldsymbol I$ and the Taylor-jet moment-matching conditions, respectively.
\end{remark}


\subsubsection{Low-rank perturbation propagation}
For a shape layer, we define
\begin{equation}
\boldsymbol W_{\pm}^{(l)}
:=\boldsymbol W^{(l)}
\pm\epsilon\boldsymbol U_t^{(l)}\boldsymbol Z_t^{(l)}
{\boldsymbol V_t^{(l)}}^{\top}.
\label{eq:det-explicit-perturbed-weight}
\end{equation}
\par
Introducing the activation matrix $\boldsymbol H$, one can get
\begin{equation}
\boldsymbol H\boldsymbol W_{\pm}^{(l)}
=\boldsymbol H\boldsymbol W^{(l)}
\pm\epsilon\boldsymbol H
\boldsymbol U_t^{(l)}\boldsymbol Z_t^{(l)}
{\boldsymbol V_t^{(l)}}^{\top}
=\boldsymbol H\boldsymbol W^{(l)}
\pm\epsilon
\bigl((\boldsymbol H\boldsymbol U_t^{(l)})
\boldsymbol Z_t^{(l)}\bigr)
{\boldsymbol V_t^{(l)}}^{\top}.
\label{eq:det-associative-identity}
\end{equation}
\par
Starting from the input $\boldsymbol H_{+}^{(0)}=\boldsymbol H_{-}^{(0)}$, induction over the layers shows that the recursive states in \eqref{eq:implicit} equal the activations of explicit networks with parameters $\boldsymbol\theta\pm\epsilon\mathcal Q_t\boldsymbol z_t$, up to floating-point round-off.
The $\pm$ signs will propagate through the complete network, and simply replacing $\boldsymbol H_{\pm}^{(l-1)}$ by an unperturbed activation after the first layer will lead to an algebraically non-equivalent form.

For either branch of layer $l$, the three contractions in \eqref{eq:det-associative-identity} require $\mathcal O(Bm_lr_l)+\mathcal O(Br_l^2)+\mathcal O(Br_ln_l)$ operations.
The first two contractions produce $B\times r_l$ intermediates, and the multiplication by ${\boldsymbol V_t^{(l)}}^{\top}$ produces a $B\times n_l$ preactivation perturbation.
The latter belongs to the ordinary forward-pass storage rather than the parameter-side storage budget.

The update for the given layer $l$ is expressed as
\begin{align}
\boldsymbol W_{t+1}^{(l)}
&=\boldsymbol W_t^{(l)}
-\alpha_t
\underbrace{\frac{\widetilde\ell_{\omega}(\boldsymbol\theta_t+\epsilon\mathcal Q_t\boldsymbol z_t;\boldsymbol x)-\widetilde\ell_{\omega}(\boldsymbol\theta_t-\epsilon\mathcal Q_t\boldsymbol z_t;\boldsymbol x)}{2\epsilon}}_{\widehat\delta_t}
\frac{\boldsymbol W_{+}^{(l)}-\boldsymbol W_{-}^{(l)}}{2\epsilon}
\notag\\
&=\boldsymbol W_t^{(l)}
-\alpha_t\widehat\delta_t
\frac{
\left(\boldsymbol W_t^{(l)}+\epsilon\boldsymbol U_t^{(l)}\boldsymbol Z_t^{(l)}{\boldsymbol V_t^{(l)}}^{\top}\right)
-\left(\boldsymbol W_t^{(l)}-\epsilon\boldsymbol U_t^{(l)}\boldsymbol Z_t^{(l)}{\boldsymbol V_t^{(l)}}^{\top}\right)
}{2\epsilon}
\notag\\
&=
\boldsymbol W_t^{(l)}
-\alpha_t\widehat\delta_t
\boldsymbol U_t^{(l)}\boldsymbol Z_t^{(l)}
{\boldsymbol V_t^{(l)}}^{\top}.
\label{eq:det-full-lowrank-update}
\end{align}
\par
We divide the row indices into separate blocks $S_1,\ldots,S_K$.
Selecting the index set $S_k$ on both sides of \eqref{eq:det-full-lowrank-update} gives
\begin{equation}
\boldsymbol W_{t+1}^{(l)}[S_k,:]
=\boldsymbol W_t^{(l)}[S_k,:]
-\alpha_t\widehat\delta_t
\boldsymbol U_t^{(l)}[S_k,:]
\left(\boldsymbol Z_t^{(l)}
{\boldsymbol V_t^{(l)}}^{\top}\right).
\label{eq:det-block-update}
\end{equation}
\par

We define the $c_{l}$ which is referred to the row number of the $S_k$, and the update in \eqref{eq:det-block-update} costs the temporary storage $\mathcal O(r_ln_l+c_{l}n_l)$, indicating that the introduction of $c_{l}$ can further reduce memory storage, while the base matrix $\boldsymbol W_t^{(l)}$ remains of the size $\mathcal O(m_ln_l)$.

Meanwhile, the biases can be included by a uniform augmentation:
\begin{equation}
\overline{\boldsymbol H}^{(l-1)}
=[\boldsymbol H^{(l-1)},\boldsymbol 1],
\qquad
\overline{\boldsymbol W}^{(l)}
=\begin{bmatrix}
\boldsymbol W^{(l)}\\ {\boldsymbol b^{(l)}}^{\top}
\end{bmatrix}.
\label{eq:det-homogeneous-bias}
\end{equation}
\par
With the homogeneous augmentation \eqref{eq:det-homogeneous-bias}, the matrix-free evaluation in \eqref{eq:det-explicit-perturbed-weight} applies the low-rank perturbation in \eqref{eq:det-full-lowrank-update}--\eqref{eq:det-block-update} directly to $\overline{\boldsymbol W}^{(l)}$, avoiding materialization of the positive and negative perturbed augmented weight matrices.


\section{Theoretical analysis}
\label{sec:theory}

The analysis considers three stochastic effects: (i) the spatial randomness in the stochastic residual, (ii) the Gaussian direction randomness in the zeroth-order estimator, and (iii) the partial coverage of the training parameter space through layer-wise subspaces.
The proposed CRNS controls the first effect, whereas $q$ and $r_l^2/(m_ln_l)$ govern the latter two, respectively.

\subsection{Mean-square regularity and CRNS stabilization}

We write the stochastic training loss as
\begin{equation}
\widetilde\ell_{\omega}(\boldsymbol\theta)
=\mathcal L(\boldsymbol\theta)+\eta_{\omega}(\boldsymbol\theta),
\qquad
\mathbb{E}_{\omega}[\eta_{\omega}(\boldsymbol\vartheta)]=0
\quad\text{for all }\boldsymbol\vartheta\in\mathcal N,
\label{eq:det-noise-field}
\end{equation}
Assume continuously differentiable sample paths on $\mathcal N$, with $\mathcal N=\mathbb R^P$ or $\boldsymbol\theta+s\epsilon\mathcal Q\boldsymbol z\in\mathcal N$ for all $s\in[-1,1]$ almost surely.
Impose the mean-square regularity condition
\begin{equation}
\sup_{\boldsymbol\vartheta\in\mathcal N}
\mathbb{E}_{\omega}[\|\nabla\eta_{\omega}(\boldsymbol\vartheta)\|_2^2]
\leq L_{\eta}^2.
\label{eq:det-ms-gradient-bound}
\end{equation}
\par
For SDGD, this follows from bounded per-term residual gradients, whereas HTE and STDE require sufficiently high probe moments.
Then \eqref{eq:det-ms-lipschitz} follows by line integration and Jensen's inequality.
\begin{equation}
\mathbb{E}_{\omega}[|\eta_{\omega}(\boldsymbol\vartheta_1)
-\eta_{\omega}(\boldsymbol\vartheta_2)|^2]
\leq L_{\eta}^2
\|\boldsymbol\vartheta_1-\boldsymbol\vartheta_2\|_2^2.
\label{eq:det-ms-lipschitz}
\end{equation}
\par
For a fixed direction $\boldsymbol p$, define the deterministic symmetric directional difference
\begin{equation}
d_{\epsilon}(\boldsymbol\theta;\boldsymbol p)
:=\frac{\mathcal L(\boldsymbol\theta+\epsilon\boldsymbol p)
-\mathcal L(\boldsymbol\theta-\epsilon\boldsymbol p)}{2\epsilon}.
\label{eq:det-d-epsilon}
\end{equation}
\par
The independent-state and CRNS estimators approximating $d_{\epsilon}(\boldsymbol\theta;\boldsymbol p)$ are
\[
\widehat\delta_{\mathrm{ind}}:=\frac{\widetilde\ell_{\omega^+}(\boldsymbol\theta+\epsilon\boldsymbol p)-\widetilde\ell_{\omega^-}(\boldsymbol\theta-\epsilon\boldsymbol p)}{2\epsilon},\qquad
\widehat\delta_{\mathrm{crn}}:=\frac{\widetilde\ell_{\omega}(\boldsymbol\theta+\epsilon\boldsymbol p)-\widetilde\ell_{\omega}(\boldsymbol\theta-\epsilon\boldsymbol p)}{2\epsilon}.
\]

\begin{trivlist}
\refstepcounter{prop}\label{prop:crns}
\item[]
\textbf{Proposition~\theprop (exact role of CRNS).}
Under \eqref{eq:det-noise-field} and \eqref{eq:det-ms-lipschitz}, conditional on $\boldsymbol p$, we have
\begin{equation}
\mathbb{E}_{\omega}
[\widehat\delta_{\mathrm{crn}}\mid\boldsymbol p]
 =d_{\epsilon}(\boldsymbol\theta;\boldsymbol p),\qquad
\operatorname{Var}_{\omega}
(\widehat\delta_{\mathrm{crn}}\mid\boldsymbol p)
\leq L_{\eta}^2\|\boldsymbol p\|_2^2.
\label{eq:det-crn-moments}
\end{equation}
\par
If $\sigma_{\eta}^2(\boldsymbol\vartheta) :=\operatorname{Var}_{\omega}[ \eta_{\omega}(\boldsymbol\vartheta)]$ is continuous at $\boldsymbol\theta$, then
\begin{equation}
\lim_{\epsilon\rightarrow0}
\epsilon^2
\operatorname{Var}_{\omega^+,\omega^-}
(\widehat\delta_{\mathrm{ind}}\mid\boldsymbol p)
=\frac12\sigma_{\eta}^2(\boldsymbol\theta).
\label{eq:det-independent-limit}
\end{equation}
\par
Together, \eqref{eq:det-crn-moments} and \eqref{eq:det-independent-limit} show that CRNS removes the $\epsilon^{-2}$ singularity while retaining finite gradient noise.
\end{trivlist}


\subsection{Pseudo-gradient moments}

Assume that the Hessian of $\mathcal L$ is $\rho$-Lipschitz on $\mathcal N$, namely,
\begin{equation}
\|\nabla^2\mathcal L(\boldsymbol\vartheta_1)
-\nabla^2\mathcal L(\boldsymbol\vartheta_2)\|_{\mathrm{op}}
\leq\rho
\|\boldsymbol\vartheta_1-\boldsymbol\vartheta_2\|_2,
\qquad
\boldsymbol\vartheta_1,\boldsymbol\vartheta_2\in\mathcal N.
\label{eq:det-hessian-lipschitz}
\end{equation}
\par
Under \eqref{eq:det-hessian-lipschitz}, the symmetric difference has the $\mathcal O(\epsilon^2)$ bias in \eqref{eq:det-central-bias}:
\begin{equation}
\left|
 d_{\epsilon}(\boldsymbol\theta;\boldsymbol p)
-\langle\nabla\mathcal L(\boldsymbol\theta),\boldsymbol p\rangle
\right|
\leq\frac{\rho\epsilon^2}{6}\|\boldsymbol p\|_2^3.
\label{eq:det-central-bias}
\end{equation}

For $\boldsymbol p=\mathcal Q\boldsymbol z$, conditional on $\mathcal Q$, let $\boldsymbol z\sim\mathcal N(\boldsymbol0,\boldsymbol I_q)$ and define
\begin{equation}
\boldsymbol a:=\mathcal Q^{\top}
\nabla\mathcal L(\boldsymbol\theta),
\qquad
\widehat{\boldsymbol g}_{\epsilon}
:=\widehat\delta_{\mathrm{crn}}\mathcal Q\boldsymbol z.
\label{eq:det-pseudogradient}
\end{equation}

Standard Gaussian moment identities, given in the supplementary material, yield the constants in Proposition~\ref{prop:pseudogradient-moments}:
\noeqref{eq:det-BV}
\begin{equation}
B_{\epsilon}
:=\frac{\rho\epsilon^2}{6}q(q+2),
\qquad
V_{\epsilon,\eta}
:=L_{\eta}^2q(q+2)
+\frac{\rho^2\epsilon^4}{18}
q(q+2)(q+4)(q+6).
\label{eq:det-BV}
\end{equation}
where $B_{\epsilon}$ bounds the central-difference bias.
The quantity $V_{\epsilon,\eta}$ collects the variance contributions of stochastic noise and finite-difference truncation.

\begin{trivlist}
\refstepcounter{prop}\label{prop:pseudogradient-moments}
\item[]
\textbf{Proposition~\theprop (projected pseudo-gradient moments).}
Under \eqref{eq:det-noise-field}, \eqref{eq:det-ms-gradient-bound}, and \eqref{eq:det-hessian-lipschitz}, conditional on a fixed basis $\mathcal Q$, the following bounds control the deviation of the conditional mean from the projected gradient and the conditional second moment of the CRNS pseudo-gradient:
\begin{align}
&\left\|
\mathbb{E}_{\omega,\boldsymbol z}
[\widehat{\boldsymbol g}_{\epsilon}\mid\mathcal Q]
-\mathcal Q\mathcal Q^{\top}
\nabla\mathcal L(\boldsymbol\theta)
\right\|_2
\leq B_{\epsilon},
\label{eq:det-estimator-bias}\\
&\mathbb{E}_{\omega,\boldsymbol z}
[\|\widehat{\boldsymbol g}_{\epsilon}\|_2^2
 \mid\mathcal Q]
\leq
2(q+2)\|\boldsymbol a\|_2^2
+V_{\epsilon,\eta}.
\label{eq:det-estimator-second}
\end{align}
\par
\eqref{eq:det-estimator-bias}--\eqref{eq:det-estimator-second} quantify the finite-radius bias and second moment.
In the deterministic zero-radius limit, the pseudo-gradient admits exact covariance, alignment, and multi-query variance identities; these are given in the supplementary material.
\end{trivlist}

\subsection{Layer-wise coverage}

Let $\boldsymbol U_l\in\mathbb R^{m_l\times r_l}$ and $\boldsymbol V_l\in\mathbb R^{n_l\times r_l}$ be independent Haar Stiefel factors, and set $\mathcal Q_l=\boldsymbol V_l\otimes\boldsymbol U_l$.
For the block-diagonal global basis and $\boldsymbol g=(\boldsymbol g_1,\ldots,\boldsymbol g_L)$, orthogonal invariance and independence give
\noeqref{eq:det-global-coverage}
\begin{equation}
\mathbb{E}[\|\mathcal Q^{\top}\boldsymbol g\|_2^2]
=\sum_{l=1}^{L}
\frac{r_l^2}{m_ln_l}\|\boldsymbol g_l\|_2^2.
\label{eq:det-global-coverage}
\end{equation}
Thus $\kappa:=\min_l r_l^2/(m_ln_l)$ lower-bounds the expected fraction of the squared gradient captured by the active subspace.
With $p_l=m_ln_l$ and $P=\sum_l p_l$, let $q_{\mathrm{tar}}$ be a nominal rank-squared budget and temporarily relax $q_l=r_l^2$ to a nonnegative real allocation.
The continuous max--min relaxation under $\sum_lq_l=q_{\mathrm{tar}}$ yields
\noeqref{eq:det-rank-solution}
\begin{equation}
q_l^{*}=q_{\mathrm{tar}}\frac{p_l}{P},
\qquad
\kappa^{*}=\frac{q_{\mathrm{tar}}}{P}.
\label{eq:det-rank-solution}
\end{equation}
The implementation instead uses integer ranks and reports
\[
q=q_{\mathrm{actual}}:=\sum_l r_l^2,
\qquad
\kappa=\kappa_{\mathrm{actual}}:=\min_l\frac{r_l^2}{m_ln_l}.
\]
Thus the continuous value $q_{\mathrm{tar}}/P$ is not asserted for the implemented ranks.
The continuous derivation, rank-capped extension, and a feasible integer-rank adjustment are given in the supplementary material.

\subsection{Projected stationarity}

Let $\{\mathcal F_t\}$ be the filtration such that $\mathcal F_t$ is generated by the iterations and all random variables sampled before iteration $t$.
The basis $\mathcal Q_t$ is chosen before $(\boldsymbol z_t,\omega_t)$ and may be reused from a previous iteration.
Conditional on $(\mathcal F_t,\mathcal Q_t)$, assume that $(\boldsymbol z_t,\omega_t)$ has the product law $\mathcal N(\boldsymbol0,\boldsymbol I_q)\otimes\mathbb P_s$.
Assume that $\mathcal L$ is bounded below by $\mathcal L_{\inf}$ and is $\beta$-smooth for $\beta>0$:
\noeqref{eq:det-smoothness}
\begin{equation}
\mathcal L(\boldsymbol y)
\leq\mathcal L(\boldsymbol x)
+\langle\nabla\mathcal L(\boldsymbol x),
\boldsymbol y-\boldsymbol x\rangle
+\frac\beta2\|\boldsymbol y-\boldsymbol x\|_2^2.
\label{eq:det-smoothness}
\end{equation}
\par
\begin{trivlist}\item[]
\refstepcounter{theorem}\label{thm:projected-stationarity}
\textbf{Theorem~\thetheorem (finite-time projected-stationarity bound).}
Suppose that the hypotheses of Proposition~\ref{prop:pseudogradient-moments} hold conditionally at every iteration with constant $\epsilon$ and $L_{\eta}$.
For the update $\boldsymbol\theta_{t+1}=\boldsymbol\theta_t-\alpha\widehat{\boldsymbol g}_t$ and step size
\noeqref{eq:det-step-condition}
\begin{equation}
0<\alpha\leq\frac{1}{4\beta(q+2)},
\label{eq:det-step-condition}
\end{equation}
one has
\begin{equation}
\frac1T\sum_{t=0}^{T-1}
\mathbb{E}[\|\mathcal Q_t^{\top}
\nabla\mathcal L(\boldsymbol\theta_t)\|_2^2]
\leq
\frac{4\Delta_0}{\alpha T}
+2B_{\epsilon}^2
+2\beta\alpha V_{\epsilon,\eta}.
\label{eq:det-projected-constant}
\end{equation}
where $\Delta_0:=\mathcal L(\boldsymbol\theta_0)-\mathcal L_{\inf}$.
The result permits arbitrary basis reuse and controls the time-averaged projected gradient.
\end{trivlist}

A variable-step and weighted-random-iterate extension is given in the supplementary material.

\subsection{Full-gradient bounds for fresh and lazy subspaces}

\eqref{eq:det-projected-constant} bounds the component of the gradient in the subspace spanned by the columns of $\mathcal Q_t$.
To obtain a bound for the full gradient norm, we require a coverage condition that quantifies how the sampled subspace represents the full parameter space.

\begin{trivlist}\item[]
\refstepcounter{cor}\label{cor:fresh-conditional-coverage}
\textbf{Corollary~\thecor (fresh conditional coverage).}
Suppose that $\mathcal Q_t$ is freshly sampled after a fixed $\mathcal F_t$ and satisfies
\begin{equation}
\mathbb{E}[\mathcal Q_t\mathcal Q_t^{\top}\mid\mathcal F_t]
\succeq\kappa\boldsymbol I_P,
\qquad \kappa>0.
\label{eq:det-fresh-coverage}
\end{equation}
\par
Combining \eqref{eq:det-fresh-coverage} with Theorem~\ref{thm:projected-stationarity} gives
\begin{equation}
\frac1T\sum_{t=0}^{T-1}
\mathbb{E}[\|\nabla\mathcal L(\boldsymbol\theta_t)\|_2^2]
\leq
\frac{4\Delta_0}{\kappa\alpha T}
+\frac{2B_{\epsilon}^2}{\kappa}
+\frac{2\beta\alpha V_{\epsilon,\eta}}{\kappa}.
\label{eq:det-full-fresh-bound}
\end{equation}
\par
For fresh Haar factors, $\kappa=\min_l r_l^2/(m_ln_l)$.
The value $q_{\mathrm{tar}}/P$ applies only when the balanced continuous allocation is exactly realized by feasible integer squares; otherwise \eqref{eq:det-full-fresh-bound} uses $q=q_{\mathrm{actual}}$ and $\kappa=\kappa_{\mathrm{actual}}$.
\end{trivlist}

For lazy refresh, let $s(t):=F\lfloor t/F\rfloor$, assume the blockwise coverage condition \eqref{eq:det-block-coverage}, and, for constant $\alpha$, $\epsilon$, and $L_{\eta}$, define $\mathcal R_T$ and $c_{\kappa}$ by \eqref{eq:det-RT}, where $\mathcal R_T$ collects initialization, finite-difference bias, and variance contributions:
\begin{equation}
\mathbb{E}[\mathcal Q_{s}\mathcal Q_{s}^{\top}\mid\mathcal F_s]\succeq\kappa\boldsymbol I_P,\qquad \mathcal Q_t=\mathcal Q_s\quad(s\leq t<s+F).
\label{eq:det-block-coverage}
\end{equation}
\begin{equation}
\mathcal R_T
:=\Delta_0
+\frac{\alpha T}{2}B_{\epsilon}^2
+\frac{\beta\alpha^2T}{2}V_{\epsilon,\eta},
\qquad
c_{\kappa}:=1+\frac\kappa2.
\label{eq:det-RT}
\end{equation}

\begin{trivlist}\item[]
\refstepcounter{theorem}\label{thm:lazy-refresh}
\textbf{Theorem~\thetheorem (full-space lazy-refresh bound with drift).}
Under the assumptions of Theorem~\ref{thm:projected-stationarity} and \eqref{eq:det-block-coverage},
\noeqref{eq:det-lazy-full-bound}
\begin{align}
\frac1T\sum_{t=0}^{T-1}
\mathbb{E}[\|\nabla\mathcal L(\boldsymbol\theta_t)\|_2^2]
\leq
\frac{16\mathcal R_T}{\kappa\alpha T}
{}+
\frac{4c_{\kappa}\beta^2F^2\alpha}{\kappa T}
\left[8(q+2)\mathcal R_T
+\alpha T V_{\epsilon,\eta}\right],
\label{eq:det-lazy-full-bound}
\end{align}
where the first term on the right side is the fresh-coverage contribution, and the second term is the price of reusing one basis while the gradient moves inside a block.
\end{trivlist}

With $\alpha=\Theta(T^{-1/2})$ and $\epsilon=\Theta(T^{-1/4})$, the fresh contribution is $\mathcal O((\kappa\sqrt T)^{-1})$, the squared finite-difference bias is $\mathcal O(T^{-1})$, and lazy refresh contributes $\mathcal O(F^2/(\kappa T))$; fixed $F$ therefore preserves the leading rate.
Complete proofs are given in the supplementary material.

\fi
\ifdefined\SDZEProofs

\section{Detailed Proofs}
\label{sec:det-proofs}

This supplementary material records the probabilistic and algebraic details underlying Section~\ref{sec:theory}.
Each derivation is presented in a form that can be verified independently.

\subsection{Indicator identities for simple random sampling}
\label{sec:proof-indicator-identities}

Let $I$ be uniformly distributed over all $b$-element subsets of $\{1,\ldots,N\}$, and let
\begin{equation}
\delta_i:=\boldsymbol 1_{\{i\in I\}}.
\label{eq:det-indicator-def}
\end{equation}
\par
Because exactly $b$ indices are selected, $\sum_{i=1}^{N}\delta_i=b$ almost surely.
By exchangeability, all $\mathbb{E}[\delta_i]$ are equal.
Taking expectations in this identity, and counting the $\binom{N-2}{b-2}$ subsets containing $i\neq j$, give
\begin{align}
N\mathbb{E}[\delta_i]&=b,
\qquad
\mathbb{E}[\delta_i]=\frac bN.
\label{eq:det-indicator-first}
\\
\mathbb{E}[\delta_i\delta_j]
&=\frac{\binom{N-2}{b-2}}{\binom Nb}
=\frac{b(b-1)}{N(N-1)}.
\label{eq:det-indicator-second}
\end{align}
\par
Since $\delta_i^2=\delta_i$,
\begin{align}
\operatorname{Var}(\delta_i)
&=\frac bN\left(1-\frac bN\right),
\label{eq:det-indicator-var}\\
\operatorname{Cov}(\delta_i,\delta_j)
&=\frac{b(b-1)}{N(N-1)}-\frac{b^2}{N^2}
=-\frac{b(N-b)}{N^2(N-1)},
\qquad i\neq j.
\label{eq:det-indicator-cov}
\end{align}
\par
The negative covariance is the algebraic source of the finite-population correction.

\subsection{Unbiased stochastic operator and cross-sampled loss}
\label{sec:proof-unbiased-loss}

Using \eqref{eq:det-random-residual} together with the indicator identities above, the sampled operator can be written as
\begin{equation}
\widetilde{\boldsymbol A}_I
=\frac Nb\sum_{i=1}^{N}\delta_i\boldsymbol a_i.
\label{eq:det-random-op-indicators}
\end{equation}
\par
Taking expectations in \eqref{eq:det-random-op-indicators} and using \eqref{eq:det-indicator-first} yields
\begin{align}
\mathbb{E}_I[\widetilde{\boldsymbol A}_I]
&=\frac Nb\sum_{i=1}^{N}
\mathbb{E}[\delta_i]\,\boldsymbol a_i
=\frac Nb\sum_{i=1}^{N}\frac bN\boldsymbol a_i
=\sum_{i=1}^{N}\boldsymbol a_i
=\boldsymbol A.
\label{eq:det-op-unbiased-proof}
\end{align}
\par
Therefore $\mathbb{E}_I[\boldsymbol r_I]=\boldsymbol r$.

For the same-sample bias, write $\boldsymbol r_I=\boldsymbol r+ (\widetilde{\boldsymbol A}_I-\boldsymbol A)$.
Expanding the squared norm,
\begin{align}
\|\boldsymbol r_I\|_2^2
&=\|\boldsymbol r\|_2^2
+2\boldsymbol r^{\top}
(\widetilde{\boldsymbol A}_I-\boldsymbol A)
+\|\widetilde{\boldsymbol A}_I-\boldsymbol A\|_2^2.
\label{eq:det-single-loss-expansion}
\end{align}
\par
By \eqref{eq:det-op-unbiased-proof}, the cross term has zero expectation.
For any zero-mean vector $\boldsymbol X$, $\mathbb{E}[\|\boldsymbol X\|_2^2] =\operatorname{tr}\mathbb{E}[\boldsymbol X\boldsymbol X^{\top}] =\operatorname{tr}\operatorname{Cov}(\boldsymbol X)$.
Hence
\begin{equation}
\mathbb{E}[\|\boldsymbol r_I\|_2^2]
=\|\boldsymbol r\|_2^2
+\operatorname{tr}
\operatorname{Cov}(\widetilde{\boldsymbol A}_I),
\label{eq:det-single-bias-proof}
\end{equation}
which is \eqref{eq:det-single-sample-loss-bias} after division by two.

For independent sets $I$ and $J$, conditioning on $I$ yields
\begin{align}
\mathbb{E}_{I,J}[
\boldsymbol r_I^{\top}\boldsymbol r_J]
&=\mathbb{E}_I\left[
\boldsymbol r_I^{\top}
\mathbb{E}_J[\boldsymbol r_J\mid I]
\right]
=\mathbb{E}_I\left[
\boldsymbol r_I^{\top}\boldsymbol r\right]
=(\mathbb{E}_I[\boldsymbol r_I])^{\top}\boldsymbol r
=\boldsymbol r^{\top}\boldsymbol r.
\label{eq:det-double-loss-proof}
\end{align}
\par
The second equality uses independence: conditional on $I$, the distribution of $J$ is unchanged.
This proves \eqref{eq:det-double-sampling-unbiased}.

\subsection{Differentiation under expectation}
\label{sec:proof-differentiation-expectation}

For the coordinate difference quotient $Q_h$, differentiability gives $Q_h\to\partial_k\widetilde\ell_{\omega}(\boldsymbol\theta)$ almost surely, while the mean-value theorem and \eqref{eq:det-dominated-gradient} give $|Q_h|\leq G(\omega)$.
Dominated convergence therefore yields
\begin{align}
\partial_k\mathbb{E}_{\omega}[
\widetilde\ell_{\omega}(\boldsymbol\theta)]
&=\lim_{h\to0}
\mathbb{E}_{\omega}[Q_h]
=\mathbb{E}_{\omega}[
\lim_{h\to0}Q_h]
=\mathbb{E}_{\omega}[
\partial_k\widetilde\ell_{\omega}(\boldsymbol\theta)].
\label{eq:det-diff-expect-coordinate}
\end{align}
\par
Applying this identity coordinatewise gives \eqref{eq:det-gradient-interchange}.

\subsection{Finite-population covariance}
\label{sec:proof-finite-pop-covariance}

Let $\boldsymbol b_i:=\boldsymbol a_i-\overline{\boldsymbol a}$, so that $\sum_i\boldsymbol b_i=\boldsymbol0$.
The indicator identities in The supplementary material gives $\widetilde{\boldsymbol A}_I-\boldsymbol A =(N/b)\sum_i\delta_i\boldsymbol b_i$ and $\sum_{i\neq j}\boldsymbol b_i\boldsymbol b_j^{\top} =-\sum_i\boldsymbol b_i\boldsymbol b_i^{\top}$.
Hence
\begin{align}
\operatorname{Cov}(\widetilde{\boldsymbol A}_I)
&=\frac{N^2}{b^2}
\left[
\frac bN-\frac{b(b-1)}{N(N-1)}
\right]
\sum_i\boldsymbol b_i\boldsymbol b_i^{\top}
\notag\\
&=\frac{N^2}{b^2}
\frac bN\frac{N-b}{N-1}
\sum_i\boldsymbol b_i\boldsymbol b_i^{\top}
\notag\\
&=\frac{N(N-b)}{b}
\frac{1}{N-1}
\sum_i\boldsymbol b_i\boldsymbol b_i^{\top}
\notag\\
&=\frac{N^2}{b}
\left(1-\frac bN\right)\boldsymbol\Sigma_a,
\label{eq:det-cov-final}
\end{align}
which is \eqref{eq:det-finite-pop-cov}.

\subsection{Matrix-free associative identity}
\label{sec:proof-matrix-free-associative}

Both recursions start from $\boldsymbol X$.
If they agree at layer $l-1$, distributivity gives
\begin{align}
\boldsymbol S_{\pm,\mathrm{exp}}^{(l)}
&=\boldsymbol H_{\pm}^{(l-1)}
\left(\boldsymbol W^{(l)}
\pm\epsilon\boldsymbol U_t^{(l)}\boldsymbol Z_t^{(l)}
{\boldsymbol V_t^{(l)}}^{\top}\right)
\notag\\
&=\boldsymbol H_{\pm}^{(l-1)}\boldsymbol W^{(l)}
\pm\epsilon
\left(
(\boldsymbol H_{\pm}^{(l-1)}\boldsymbol U_t^{(l)})
\boldsymbol Z_t^{(l)}
\right){\boldsymbol V_t^{(l)}}^{\top}
\notag\\
&=\boldsymbol S_{\pm,\mathrm{imp}}^{(l)}.
\label{eq:det-assoc-induction-step}
\end{align}
\par
Applying $\sigma$ completes the induction without materializing a dense perturbation.
For a row selector $\boldsymbol E_{S_k}$,
\begin{align}
\boldsymbol E_{S_k}\boldsymbol W_{t+1}^{(l)}
&=\boldsymbol E_{S_k}\boldsymbol W_t^{(l)}
-\alpha_t\widehat\delta_t
(\boldsymbol E_{S_k}\boldsymbol U_t^{(l)})
\boldsymbol Z_t^{(l)}{\boldsymbol V_t^{(l)}}^{\top},
\label{eq:det-block-selector}
\end{align}
which is \eqref{eq:det-block-update}; the partitioned selectors recover the full update.

\subsection{Mean-square Lipschitz continuity}

Let $\boldsymbol h:=\boldsymbol\vartheta_1-\boldsymbol\vartheta_2$ and $\boldsymbol\gamma(s):=\boldsymbol\vartheta_2+s\boldsymbol h$.
The fundamental theorem of calculus, Jensen's inequality, and Tonelli's theorem give
\begin{align}
\mathbb{E}_{\omega}\left[
|\eta_{\omega}(\boldsymbol\vartheta_1)
-\eta_{\omega}(\boldsymbol\vartheta_2)|^2\right]
&\leq\|\boldsymbol h\|_2^2\int_0^1
\mathbb{E}_{\omega}\left[
\|\nabla\eta_{\omega}(\boldsymbol\gamma(s))\|_2^2\right]\,ds
\leq L_{\eta}^2\|\boldsymbol h\|_2^2.
\label{eq:det-noise-ms-final}
\end{align}
\par
This proves \eqref{eq:det-ms-lipschitz}.

\subsection{Complete proof of Proposition~\ref{prop:crns}}

Substituting \eqref{eq:det-noise-field} into the CRNS estimator, and then taking conditional expectation using the zero mean of $\eta_{\omega}$, give
\begin{align}
\widehat\delta_{\mathrm{crn}}
&=d_{\epsilon}(\boldsymbol\theta;\boldsymbol p)
+\frac{
\eta_{\omega}(\boldsymbol\theta+\epsilon\boldsymbol p)
-\eta_{\omega}(\boldsymbol\theta-\epsilon\boldsymbol p)}{2\epsilon}.
\label{eq:det-crn-decomposition}\\
\mathbb{E}_{\omega}[
\widehat\delta_{\mathrm{crn}}\mid\boldsymbol p]
&=d_{\epsilon}(\boldsymbol\theta;\boldsymbol p)
+\frac{0-0}{2\epsilon}
=d_{\epsilon}(\boldsymbol\theta;\boldsymbol p),
\label{eq:det-crn-mean-proof}
\end{align}
which proves \eqref{eq:det-crn-moments}.

Define the noise quotient
\begin{equation}
X_{\epsilon}
:=\frac{
\eta_{\omega}(\boldsymbol\theta+\epsilon\boldsymbol p)
-\eta_{\omega}(\boldsymbol\theta-\epsilon\boldsymbol p)}{2\epsilon}.
\label{eq:det-Xepsilon}
\end{equation}
\par
Since $\mathbb{E}_{\omega}[X_{\epsilon}]=0$, the conditional variance is
\begin{align}
\operatorname{Var}_{\omega}
(\widehat\delta_{\mathrm{crn}}\mid\boldsymbol p)
&=\mathbb{E}_{\omega}[|X_{\epsilon}|^2]
\notag\\
&=\frac{1}{4\epsilon^2}
\mathbb{E}_{\omega}\left[
|\eta_{\omega}(\boldsymbol\theta+\epsilon\boldsymbol p)
-\eta_{\omega}(\boldsymbol\theta-\epsilon\boldsymbol p)|^2\right]
\notag\\
&\leq\frac{L_{\eta}^2}{4\epsilon^2}
\|2\epsilon\boldsymbol p\|_2^2
=L_{\eta}^2\|\boldsymbol p\|_2^2,
\label{eq:det-crn-var-proof}
\end{align}
where \eqref{eq:det-ms-lipschitz} was used with the two perturbed parameters.
This proves \eqref{eq:det-crn-moments}.

For independently sampled spatial states, the estimator is
\begin{align}
\widehat\delta_{\mathrm{ind}}
&=d_{\epsilon}(\boldsymbol\theta;\boldsymbol p)
+\frac{
\eta_{\omega^+}(\boldsymbol\theta+\epsilon\boldsymbol p)
-\eta_{\omega^-}(\boldsymbol\theta-\epsilon\boldsymbol p)}{2\epsilon}.
\label{eq:det-ind-decomposition}
\end{align}
\par
The deterministic term is constant, and the independent noise terms have zero covariance.
Therefore,
\begin{align}
\operatorname{Var}_{\omega^+,\omega^-}
(\widehat\delta_{\mathrm{ind}}\mid\boldsymbol p)
&=\frac{1}{4\epsilon^2}
\left[
\sigma_{\eta}^2(\boldsymbol\theta+\epsilon\boldsymbol p)
+\sigma_{\eta}^2(\boldsymbol\theta-\epsilon\boldsymbol p)
\right].
\label{eq:det-ind-exact-var}
\end{align}
\par
Multiplying by $\epsilon^2$ and using continuity at $\boldsymbol\theta$ gives
\begin{align}
\lim_{\epsilon\downarrow0}
\epsilon^2\operatorname{Var}
(\widehat\delta_{\mathrm{ind}}\mid\boldsymbol p)
&=\frac14
\left[\sigma_{\eta}^2(\boldsymbol\theta)
+\sigma_{\eta}^2(\boldsymbol\theta)\right]
=\frac12\sigma_{\eta}^2(\boldsymbol\theta),
\label{eq:det-ind-limit-proof}
\end{align}
which proves \eqref{eq:det-independent-limit}.

For the final assertion, mean-square differentiability provides an $L^2$ random vector $\boldsymbol G_{\omega}(\boldsymbol\theta)$ such that
\begin{equation}
\frac{
\mathbb{E}_{\omega}\left[
|\eta_{\omega}(\boldsymbol\theta+\boldsymbol h)
-\eta_{\omega}(\boldsymbol\theta)
-\langle\boldsymbol G_{\omega}(\boldsymbol\theta),
\boldsymbol h\rangle|^2\right]}
{\|\boldsymbol h\|_2^2}
\longrightarrow0
\qquad(\boldsymbol h\to\boldsymbol0).
\label{eq:det-ms-diff-def}
\end{equation}
\par
For signs $\pm$, define
\begin{equation}
R_{\pm}(\epsilon)
:=\eta_{\omega}(\boldsymbol\theta\pm\epsilon\boldsymbol p)
-\eta_{\omega}(\boldsymbol\theta)
\mp\epsilon\langle\boldsymbol G_{\omega}(\boldsymbol\theta),
\boldsymbol p\rangle,
\qquad
\|R_{\pm}(\epsilon)\|_{L^2}=o(\epsilon\|\boldsymbol p\|_2),
\label{eq:det-ms-rem-plus}
\end{equation}
by \eqref{eq:det-ms-diff-def}.
Subtracting the two expansions and applying the triangle inequality in $L^2$ give
\begin{align}
X_{\epsilon}
-\langle\boldsymbol G_{\omega}(\boldsymbol\theta),
\boldsymbol p\rangle
&=\frac{R_{+}(\epsilon)-R_{-}(\epsilon)}{2\epsilon}.
\label{eq:det-ms-sym-rem}\\
\left\|
X_{\epsilon}
-\langle\boldsymbol G_{\omega}(\boldsymbol\theta),
\boldsymbol p\rangle
\right\|_{L^2}
&\leq\frac{
\|R_{+}(\epsilon)\|_{L^2}
+\|R_{-}(\epsilon)\|_{L^2}}{2\epsilon}
\longrightarrow0,
\label{eq:det-ms-limit-proof}
\end{align}
which is \eqref{eq:det-crn-l2-limit}.

\subsection{Total-variance decomposition and the directional limit}

Let $Y:=\widehat\delta_{\mathrm{crn}}$.
Conditional on $\mathcal Q$, the law of total variance yields
\begin{equation}
\operatorname{Var}(Y\mid\mathcal Q)
=\mathbb{E}_{\boldsymbol z}[
\operatorname{Var}(Y\mid\boldsymbol z,\mathcal Q)
 \mid\mathcal Q]
+\operatorname{Var}_{\boldsymbol z}
(\mathbb{E}[Y\mid\boldsymbol z,\mathcal Q]
 \mid\mathcal Q).
\label{eq:det-total-var-law}
\end{equation}
\par
By \eqref{eq:det-crn-moments}, the inner conditional mean equals $d_{\epsilon}(\boldsymbol\theta;\mathcal Q\boldsymbol z)$, which proves \eqref{eq:det-total-variance}.

For the first term, \eqref{eq:det-crn-moments} yields
\begin{align}
\mathbb{E}_{\boldsymbol z}\left[
\operatorname{Var}_{\omega}
(\widehat\delta_{\mathrm{crn}}
 \mid\mathcal Q\boldsymbol z)\right]
&\leq L_{\eta}^2
\mathbb{E}[\|\mathcal Q\boldsymbol z\|_2^2]
\notag\\
&=L_{\eta}^2
\mathbb{E}[\boldsymbol z^{\top}
\mathcal Q^{\top}\mathcal Q\boldsymbol z]
=L_{\eta}^2\mathbb{E}[\|\boldsymbol z\|_2^2]
=L_{\eta}^2q.
\label{eq:det-total-var-first}
\end{align}
\par
Set $\boldsymbol a=\mathcal Q^{\top}\nabla\mathcal L(\boldsymbol\theta)$.
The central-difference bias bound in \eqref{eq:det-central-bias}, proved below, gives
\begin{equation}
|d_{\epsilon}(\boldsymbol\theta;\mathcal Q\boldsymbol z)
-\boldsymbol a^{\top}\boldsymbol z|
\leq\frac{\rho\epsilon^2}{6}\|\boldsymbol z\|_2^3.
\label{eq:det-d-L2-bound}
\end{equation}
\par
Squaring the bound and taking expectation yields
\begin{equation}
\mathbb{E}[|d_{\epsilon}-\boldsymbol a^{\top}\boldsymbol z|^2]
\leq\frac{\rho^2\epsilon^4}{36}
\mathbb{E}[\|\boldsymbol z\|_2^6]
\longrightarrow0.
\label{eq:det-d-L2-conv}
\end{equation}
\par
Hence $d_{\epsilon}\to\boldsymbol a^{\top}\boldsymbol z$ in $L^2$.
Because convergence in $L^2$ implies convergence of both first and second moments,
\begin{equation}
\operatorname{Var}_{\boldsymbol z}(d_{\epsilon})
\longrightarrow
\operatorname{Var}(\boldsymbol a^{\top}\boldsymbol z)
=\|\boldsymbol a\|_2^2.
\label{eq:det-directional-var-limit}
\end{equation}

\subsection{Derivation of the spatial-resource bound}

Define the three ANOVA components by
\begin{equation}
\boldsymbol X:=\frac1B\sum_{s=1}^{B}\boldsymbol\xi_s,
\quad
\boldsymbol Z:=\frac1b\sum_{j=1}^{b}\boldsymbol\zeta_j,
\quad
\boldsymbol C:=\frac1{Bb}
\sum_{s=1}^{B}\sum_{j=1}^{b}\boldsymbol\chi_{s,j}.
\label{eq:det-anova-components}
\end{equation}
\par
The assumed orthogonality of the three groups yields
\begin{equation}
\mathbb{E}[\|\boldsymbol X+\boldsymbol Z+\boldsymbol C\|_2^2]
=\mathbb{E}[\|\boldsymbol X\|_2^2]
+\mathbb{E}[\|\boldsymbol Z\|_2^2]
+\mathbb{E}[\|\boldsymbol C\|_2^2].
\label{eq:det-anova-orthogonal}
\end{equation}
\par
For the collocation component, expanding the square gives
\begin{align}
\mathbb{E}[\|\boldsymbol X\|_2^2]
&=\frac1{B^2}
\sum_{s=1}^{B}\sum_{s'=1}^{B}
\mathbb{E}[\langle\boldsymbol\xi_s,
\boldsymbol\xi_{s'}\rangle]
\notag\\
&=\frac1{B^2}
\sum_{s=1}^{B}
\mathbb{E}[\|\boldsymbol\xi_s\|_2^2]
\leq\frac{\sigma_x^2}{B},
\label{eq:det-anova-x}
\end{align}
where cross terms vanish for $s\neq s'$.
Similarly,
\begin{equation}
\mathbb{E}[\|\boldsymbol Z\|_2^2]
\leq\frac{\sigma_D^2}{b}.
\label{eq:det-anova-z}
\end{equation}
\par
For the interaction component, the same calculation yields
\begin{align}
\mathbb{E}[\|\boldsymbol C\|_2^2]
&=\frac1{B^2b^2}
\sum_{s,j}\sum_{s',j'}
\mathbb{E}[\langle\boldsymbol\chi_{s,j},
\boldsymbol\chi_{s',j'}\rangle]
\notag\\
&=\frac1{B^2b^2}
\sum_{s=1}^{B}\sum_{j=1}^{b}
\mathbb{E}[\|\boldsymbol\chi_{s,j}\|_2^2]
\leq\frac{\sigma_{xD}^2}{Bb}.
\label{eq:det-anova-c}
\end{align}
\par
Combining \eqref{eq:det-anova-orthogonal}--\eqref{eq:det-anova-c} yields \eqref{eq:det-resource-bound}.
When the operator indices are sampled without replacement, \eqref{eq:det-finite-pop-cov} multiplies the corresponding index-sampling covariance by $1-b/N$.

\fi
\let\SDZEMainTheory\undefined

\section{Numerical experiments}
\label{sec:exp}

\setcounter{bottomnumber}{2}
\renewcommand{\bottomfraction}{.35}
\renewcommand{\textfraction}{.05}
\setlength{\floatsep}{6pt plus 2pt minus 2pt}
\setlength{\intextsep}{6pt plus 2pt minus 2pt}

We evaluate SDZE on high-dimensional second-order PDEs and high-order KdV-type equations, reporting relative $L_2$ error, peak GPU memory, and iteration time.
The high-dimensional accuracy tests span Poisson, Allen--Cahn, Sine--Gordon, and HJB--LQG problems from $d=10$ to $d=50\mathrm{M}$; the extreme-memory study extends the Allen--Cahn problem to $d=100\mathrm{M}$.
HTE~\cite{hu2024hutch}, STDE~\cite{shi2024stochastic,shi2026stde++}, SDGD~\cite{hu2024sdgd}, FOBAD~\cite{griewank08_evaluat_deriv}, ForwardLap~\cite{li2024computational}, and RS-PINN~\cite{hu2025bias} provide first-order references.
SDZE combines stochastic spatial estimation with a layer-wise low-rank zeroth-order update.

\subsection{Experimental setup}
Unless otherwise noted, runnable implementations use JAX on a single NVIDIA GeForce RTX 5090 GPU.
The extreme-dimensional Allen--Cahn memory study uses two NVIDIA GeForce RTX 5090 GPUs with 64~GB total memory.
Each benchmark uses a four-layer, width-128, fully connected PINN with $\tanh$ activations.
The first-order baselines use Adam, whereas SDZE applies the direct pseudo-gradient update in \eqref{eq:delta} without parameter-wise moment states.
Unless stated otherwise, the architecture, spatial sampling budget, random seeds, iteration count, and evaluation grid are fixed across methods.
OOM denotes a run that exceeds the available memory of the corresponding hardware configuration.
Accuracy is the mean relative discrete $L_2$ error over 5 training seeds.
Iteration time is measured after warm-up.
Peak memory is the maximum allocated GPU memory for single-GPU runs and the peak aggregate allocation for the two-GPU extreme-memory study.
The SDZE implementation is publicly available at \url{https://github.com/liangzhangyong/SDZE}.

\subsection{Scalability for high-dimensional PDEs}
\label{sec:high-dim-pde}

\subsubsection{Benchmarks}
We use the following manufactured solution on the $d$-dimensional unit ball $\mathbb{B}^{d}$:
\begin{equation}
u_{\mathrm{exact}}(\vb{x})
=(1-\|\vb{x}\|_2^2)
\left(
\sum_{i=1}^{d-1} c_i
\sin\!\left(x_i+\cos(x_{i+1})+x_{i+1}\cos(x_i)\right)
\right),
\end{equation}
where $c_i\stackrel{\mathrm{i.i.d.}}{\sim}\mathcal{N}(0,1)$.
For each dimension, $c_i$, the source term, and the evaluation grid are fixed across methods.
The prefactor $1-\|\vb{x}\|_2^2$ enforces the homogeneous boundary condition on $\partial\mathbb{B}^{d}$, so that the prescribed boundary data do not reveal the exact solution.
For each operator $\mathcal{N}$, we set $f(\vb{x})=\mathcal{N}u_{\mathrm{exact}}(\vb{x})$; then the problem is transformed as
\begin{equation}
\begin{split}
\mathcal{N} u(\vb{x}) =& f(\vb{x}), \quad \vb{x} \in \mathbb{B}^{d}, \\
u(\vb{x}) =& 0, \quad\quad \vb{x} \in \partial\mathbb{B}^{d},
\end{split}
\end{equation}
where $\mathcal{N}$ is either a linear or nonlinear operator.

The three benchmarks are tested:
\begin{itemize}
  \item Poisson equation: $\mathcal{N} u(\vb{x})=\laplacian u(\vb{x}).$ 
  \item Allen--Cahn equation: $\mathcal{N} u(\vb{x})=\laplacian u(\vb{x}) + u(\vb{x}) - u(\vb{x})^{3}.$ 
  \item Sine--Gordon equation: $\mathcal{N} u(\vb{x})=\laplacian u(\vb{x}) + \sin ( u(\vb{x})).$
\end{itemize}


In addition, the HJB--LQG comparison uses the terminal-value problem
\begin{equation}
\begin{aligned}
\partial_t u(\vb{x},t)+\Delta_{\vb{x}}u(\vb{x},t)
-\|\nabla_{\vb{x}}u(\vb{x},t)\|_2^2&=0,
&& (\vb{x},t)\in\mathbb R^d\times[0,T],\\
u(\vb{x},T)&=g(\vb{x}),
&& \vb{x}\in\mathbb R^d,
\end{aligned}
\label{eq:hjb-lqg-benchmark}
\end{equation}
where $g$ denotes the terminal cost.


\subsubsection{Accuracy and resource scaling}

\begin{figure}[t]
  \centering
  \includegraphics[width=\linewidth]{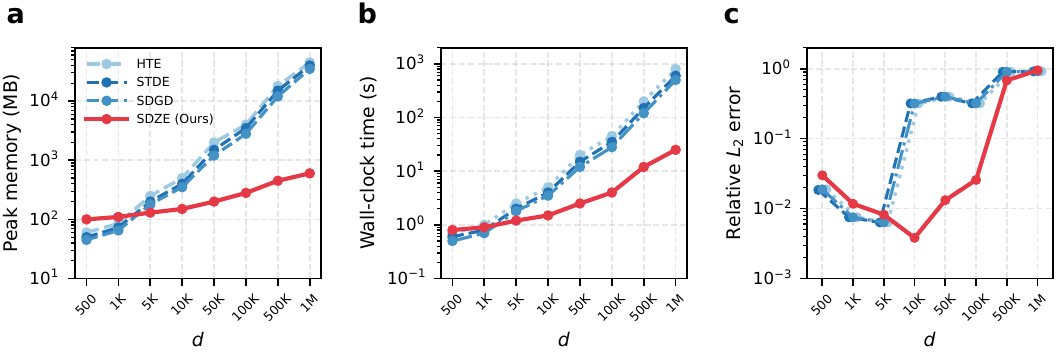}
  \vspace{-16pt}
  \caption{Peak GPU memory, training wall-clock time, and relative $L_2$ error of SDZE and baselines across increasing dimensions.}
  \label{fig:teaser}
\end{figure}

Tables~\ref{tab:allen-cahn-speed}--\ref{tab:allen-cahn-mem} quantify the iteration-time and peak-memory costs across dimensions.
Figure~\ref{fig:teaser} summarizes the overall accuracy--resource trade-off.

\begin{figure}[htbp]
  \centering
  \includegraphics[width=0.82\textwidth]{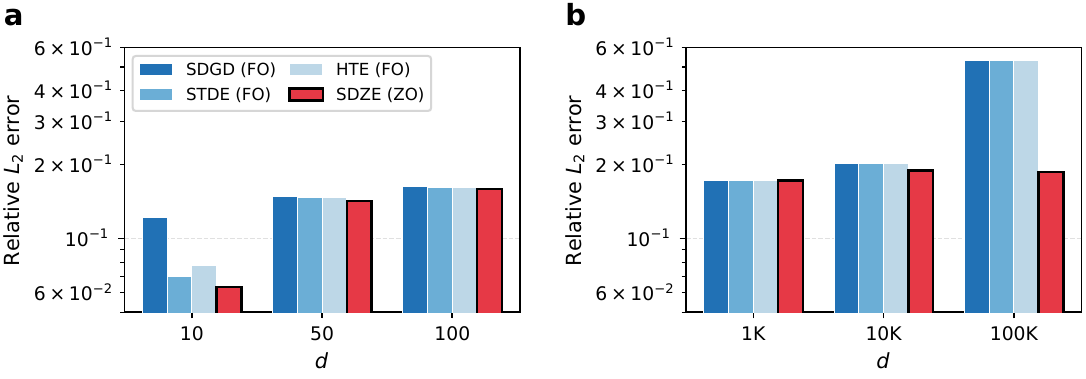}
  \vspace{-8pt}
  \caption{Relative $L_2$ errors of SDZE and baselines across dimensions for Allen--Cahn problems.}
  \label{fig:accuracy_comparison}
\end{figure}


\begin{figure}[htbp]
  \centering
  \includegraphics[width=0.88\textwidth]{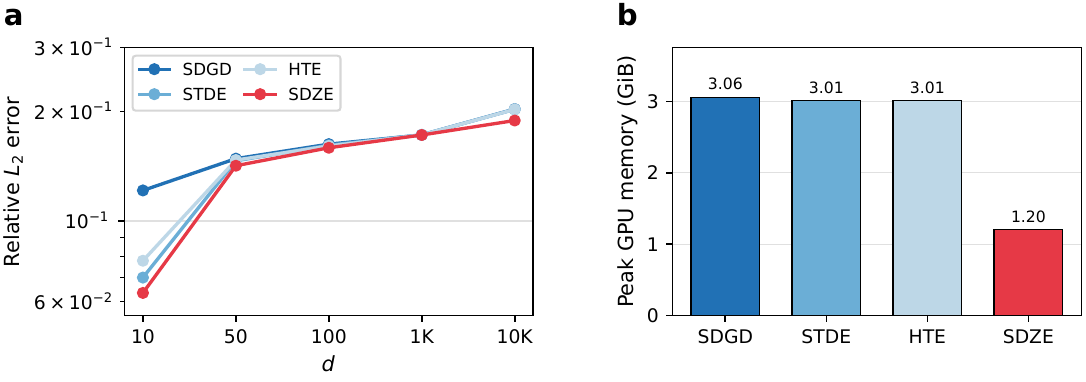}
  \vspace{-8pt}
  \caption{Relative $L_2$ errors across dimensions and peak GPU memory of the stochastic estimators.}
  \label{fig:accuracy}
\end{figure}

\begin{figure}[htbp]
  \centering
  \includegraphics[width=0.85\textwidth]{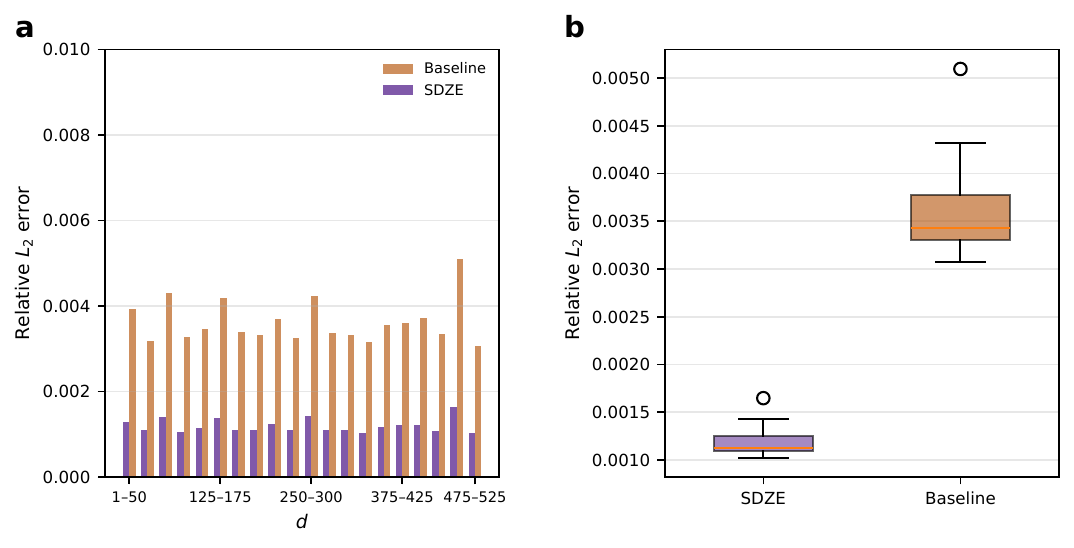}
  \vspace{-8pt}
  \caption{Relative $L_2$ errors of SDZE and existing baselines of the 1000-D Allen--Cahn equation.}
  \label{fig:1000d_slice_errors}
\end{figure}

Figures~\ref{fig:accuracy_comparison} and~\ref{fig:accuracy} show how solution accuracy evolves with dimension without a systematic loss of quality from the resource reduction.
Figure~\ref{fig:1000d_slice_errors} further shows that the improvement is accompanied by reduced variation across representative coordinate groups rather than being confined to a single global error statistic.
Across the common executable regimes, SDZE remains competitive with the stochastic first-order estimators despite restricting each parameter update to a low-dimensional random subspace.
This observation is consistent with the coverage analysis in Section~\ref{sec:theory}: individual updates access only a fraction of the parameter space, but successive randomized subspaces change the explored directions, while CRNS controls stochastic finite-difference noise within each update.

\begin{table}[htbp]
  \footnotesize
\centering
\caption{Iteration time (s/it) of SDZE and baselines for the sine-coupled Allen--Cahn problems.}
\label{tab:allen-cahn-speed}
{\renewcommand{\arraystretch}{1.15}
\resizebox{\textwidth}{!}{%
\begin{tabular}{lcccccc}
\hline
Method &  100 D & 1K D & 10K D & 100K D & 1M D & 10M D \\ \hline
Backward mode (FO) \cite{hu2024sdgd} & 8.43e-03 & 1.16e-02 & 1.87e-02 & OOM & OOM & OOM \\ \hline
Mixed-mode (AD-Spatial + ZO-Param) \cite{hu2024sdgd,liu2020primer} & 2.042e-03 & 2.215e-03 & 4.470e-03 & 2.019e-02 & 2.078e-01 & OOM \\ \hline
Parallelized backward mode (FO) \cite{hu2024sdgd,griewank08_evaluat_deriv} & 2.498e-04 & 2.846e-04 & 4.178e-04 & 2.676e-03 & 2.315e-02 & OOM \\ \hline
Forward-over-Backward (FO) \cite{hu2024sdgd,griewank08_evaluat_deriv} & 2.755e-04 & 2.898e-04 & 5.155e-04 & 2.557e-03 & 2.833e-02 & OOM \\ \hline
ForwardLap (FO) \cite{li2024computational} & 2.228e-04 & 1.193e-03 & 2.358e-02 & OOM & OOM & OOM \\ \hline
HTE (FO) \cite{hu2024hutch} & 2.663e-04 & 3.422e-04 & 4.476e-04 & 1.555e-03 & 2.249e-02 & OOM \\ \hline
RS-PINN (FO) \cite{hu2025bias} & 2.865e-04 & 6.054e-04 & 2.936e-03 & 2.892e-02 & OOM & OOM \\ \hline
STDE (FO) \cite{shi2024stochastic,shi2026stde++} & 2.492e-04 & 2.602e-04 & 3.934e-04 & 1.425e-03 & 2.263e-02 & OOM \\ \hline
SDGD (FO) \cite{hu2024sdgd} & 2.498e-04 & 2.846e-04 & 4.178e-04 & 2.676e-03 & 2.315e-02 & OOM \\ \hline
FOBAD (FO) \cite{griewank08_evaluat_deriv,hu2024sdgd} & 2.755e-04 & 2.898e-04 & 5.155e-04 & 2.557e-03 & 2.833e-02 & OOM \\ \hline
\textbf{SDZE (Ours, ZO)} & \textbf{2.113e-04} & \textbf{2.171e-04} & \textbf{2.814e-04} & \textbf{1.363e-03} & \textbf{1.518e-02} & \textbf{9.477e-01} \\ \hline
\end{tabular}}}
\end{table}

\begin{table}[htbp]
\centering
\caption{Peak memory (MB) of SDZE and baselines for the sine-coupled Allen--Cahn problems.}
\label{tab:allen-cahn-mem}
\resizebox{0.82\textwidth}{!}{%
{\renewcommand{\arraystretch}{1.15}
\begin{tabular}{lcccccc}
\hline
Method & 100 D & 1K D & 10K D & 100K D & 1M D & 10M D \\ \hline
Backward mode (FO) & 1328 & 1788 & 4527 & OOM & OOM & OOM \\ \hline
Mixed-mode (AD-Spatial + ZO-Param) & 553 & 565 & 1217 & 261 & 2575 & OOM \\ \hline
Parallelized backward mode (FO) & 539 & 579 & 1177 & 4931 & 6086  & OOM \\ \hline
Forward-over-Backward (FO) & 537 & 579 & 1519 & 4929 & 9533 & OOM \\ \hline
ForwardLap (FO) & 507 & 913 & 5505 & OOM & OOM & OOM \\ \hline
HTE (FO) & 69 & 73 & 163 & 948 & 5835 & OOM \\ \hline
RS-PINN (FO) & 130 & 262 & 1100 & 10326 & OOM & OOM \\ \hline
STDE (FO) & 69 & 73 & 137 & 719 & 5704 & OOM \\ \hline
SDGD (FO) & 54 & 65 & 176 & 796 & 6086 & OOM \\ \hline
FOBAD (FO) & 54 & 65 & 176 & 817 & 9533 & OOM \\ \hline
\textbf{SDZE (Ours, ZO)}
& \textbf{16} & \textbf{18} & \textbf{45} & \textbf{384}
& \textbf{1129} & \textbf{7109} \\ \hline
\end{tabular}}}
\end{table}

\begin{figure}[htbp]
  \centering
  \includegraphics[width=\textwidth]{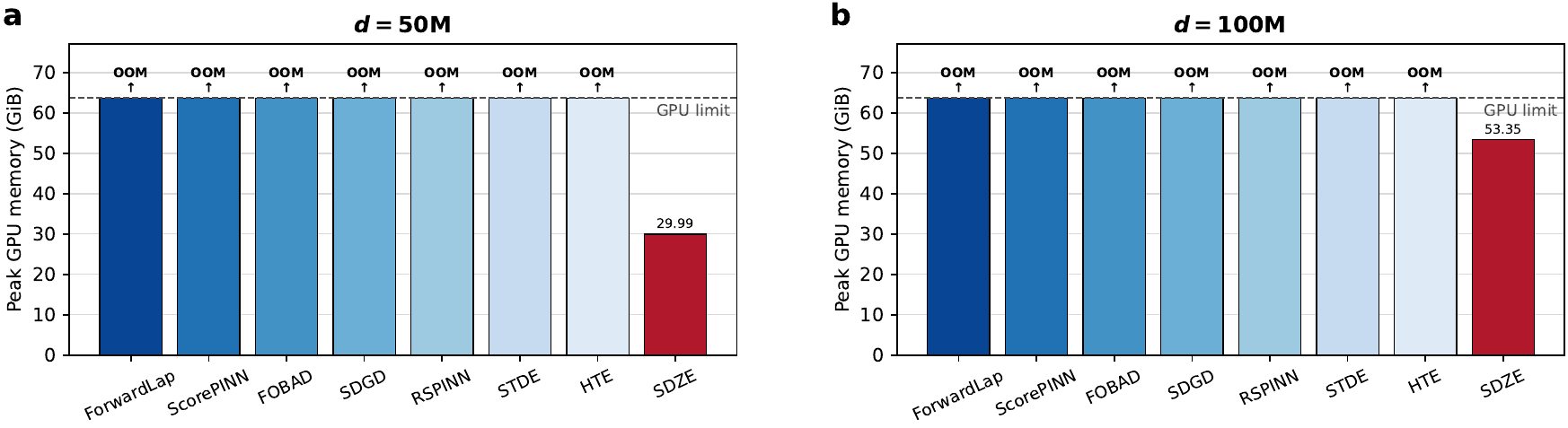}
  \vspace{-12pt}
  \caption{Memory of the compared methods for the Allen--Cahn problem at $d=50\mathrm{M}$ and $d=100\mathrm{M}$.}
  \label{fig:sota-peak-memory}
\end{figure}

\begin{figure}[htbp]
  \centering
  \includegraphics[width=0.85\textwidth]{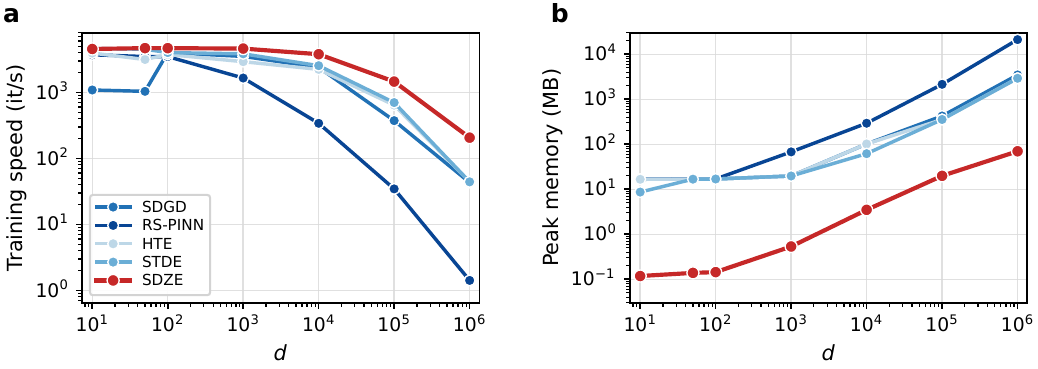}
  \vspace{-8pt}
  \caption{Training-speed and peak memory scaling for solving the Allen--Cahn equation.}
  \label{fig:sdze_scaling_corrected}
\end{figure}

Figure~\ref{fig:sota-peak-memory} extends the Allen--Cahn memory comparison to $d=50\mathrm{M}$ and $d=100\mathrm{M}$.
At $d=50\mathrm{M}$, SDZE requires 29.99~GB of aggregate GPU memory, while all compared baselines are OOM.
At $d=100\mathrm{M}$, SDZE remains executable with 53.35~GB, whereas the baselines remain OOM.
This feasibility gap is consistent with SDZE's matrix-free update, which avoids reverse-mode parameter-gradient storage.
Together with Figure~\ref{fig:sdze_scaling_corrected}, these results show that SDZE expands the executable dimension range when parameter-side memory limits the first-order baselines.

Tables~\ref{tab:multi-pde-accuracy}--\ref{tab:hjb-lqg} provide quantitative comparisons across different PDE families.
These results show that the accuracy--resource behavior persists across linear, nonlinear, and control-oriented PDEs, although its strength remains problem dependent.

\begin{table}[htbp]
\centering
\caption{Relative $L_2$ errors across different benchmarks.}
\label{tab:multi-pde-accuracy}
{\renewcommand{\arraystretch}{1.10} \setlength{\tabcolsep}{3pt}
\resizebox{0.5\textwidth}{!}{%
\begin{tabular}{llccc}
\hline
Method & $d$ & Sine--Gordon & Allen--Cahn & Poisson \\ \hline
\multirow{3}{*}{SDGD} & 10  & 7.52e-03 & 6.38e-03 & 7.69e-03 \\
                      & 50  & 8.06e-03 & 8.34e-03 & 4.69e-03 \\
                      & 100 & 7.10e-03 & 6.41e-03 & 4.80e-03 \\ \hline
\multirow{3}{*}{FOBAD}& 10  & 7.52e-03 & 6.38e-03 & 7.69e-03 \\
                      & 50  & 8.06e-03 & 8.34e-03 & 4.69e-03 \\
                      & 100 & 7.10e-03 & 6.41e-03 & 4.80e-03 \\ \hline
\multirow{3}{*}{ForwardLap}
                      & 10  & 9.63e-03 & 8.98e-03 & 9.02e-03 \\
                      & 50  & 6.92e-03 & 7.31e-03 & 1.11e-01 \\
                      & 100 & 7.05e-03 & 6.29e-03 & 4.86e-03 \\ \hline
\multirow{3}{*}{HTE}  & 10  & 4.55e-03 & 4.56e-03 & 4.52e-03 \\
                      & 50  & 7.05e-03 & 7.44e-03 & 4.02e-03 \\
                      & 100 & 7.01e-03 & 6.30e-03 & 4.83e-03 \\ \hline
\multirow{3}{*}{STDE} & 10  & 3.09e-03 & 3.54e-03 & 5.01e-03 \\
                      & 50  & 7.50e-03 & 7.79e-03 & 4.04e-03 \\
                      & 100 & 7.09e-03 & 6.38e-03 & 4.87e-03 \\ \hline
\multirow{3}{*}{\textbf{SDZE (Ours)}}
                      & 10  & \textbf{2.96e-03} & \textbf{2.46e-03} & \textbf{2.21e-03} \\
                      & 50  & \textbf{3.87e-03} & \textbf{3.21e-03} & \textbf{2.57e-03} \\
                      & 100 & \textbf{3.48e-03} & \textbf{2.96e-03} & \textbf{2.82e-03} \\ \hline
\end{tabular}}}
\end{table}

\begin{table}[htbp]
\footnotesize
\centering
\caption{Relative $L_2$ errors across dimensions for Allen--Cahn problems.}
\label{tab:allen-cahn-extreme}
{\renewcommand{\arraystretch}{1.15}
\resizebox{0.92\textwidth}{!}{%
\begin{tabular}{lccccccc}
\hline
Method & D=10 & D=100 & D=1K & D=10K & D=100K & D=1M & D=10M \\ \hline
SDGD & 2.17e-02 & 8.93e-03 & 7.35e-03 & 4.85e-03 & 3.45e-02 & 1.15e-01 & OOM \\ \hline
FOBAD & 2.17e-02 & 8.93e-03 & 7.50e-03 & 4.43e-03 & 3.84e-02 & 1.16e-01 & OOM \\ \hline
HTE & 2.09e-02 & 8.92e-03 & 7.47e-03 & 4.73e-03 & 3.38e-02 & 1.10e-01 & OOM \\ \hline
STDE & 2.06e-02 & 8.59e-03 & 7.33e-03 & 4.64e-03 & 3.98e-02 & 1.17e-01 & OOM \\ \hline
\textbf{SDZE (Ours)} & \textbf{4.07e-03} & \textbf{7.13e-03} & \textbf{4.47e-03} & \textbf{3.51e-03} & \textbf{2.63e-03} & \textbf{8.93e-03} & \textbf{1.37e-03} \\ \hline
\end{tabular}}}
\end{table}

\begin{table}[htbp]
\centering
\caption{Relative $L_2$ errors for HJB--LQG problems across dimensions.}
\label{tab:hjb-lqg}
\resizebox{0.48\textwidth}{!}{%
{\renewcommand{\arraystretch}{1.15}
\begin{tabular}{lccc}
\hline
Method & D=10 & D=50 & D=100 \\ \hline
SDGD & 1.11e-03 & 1.33e-02 & 2.14e-02 \\ \hline
FOBAD & 1.11e-03 & 1.33e-02 & 2.14e-02 \\ \hline
HTE & 1.23e-03 & 1.32e-02 & 2.02e-02 \\ \hline
STDE & 1.08e-03 & 1.30e-02 & 2.15e-02 \\ \hline
\textbf{SDZE (Ours)} & \textbf{6.40e-04} & \textbf{1.13e-02} & \textbf{1.88e-02} \\ \hline
\end{tabular}}}
\end{table}

\begin{figure}[htbp]
  \centering
  \includegraphics[width=\textwidth]{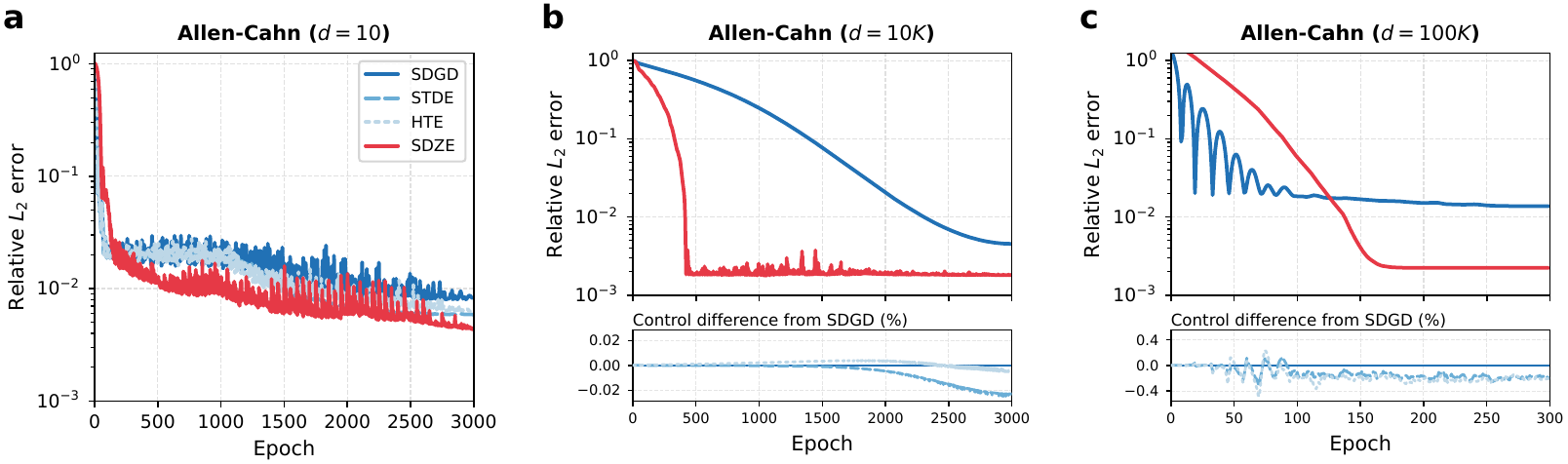}
  \vspace{-18pt}
  \caption{Relative $L_2$ error trajectories for the Allen--Cahn problems across dimensions.}
  \label{fig:convergence}
\end{figure}

Figure~\ref{fig:convergence} complements these endpoint comparisons by exposing the optimization trajectories.
The differences between methods develop over training rather than appearing only at the final iterate, and the contrast becomes more visible in the more demanding dimensional regimes.
This behavior is consistent with the growing computational and stochastic burden associated with maintaining a full first-order parameter-update path.
Together, these results suggest that SDZE's low-rank, variance-controlled updates are most favorable when parameter-gradient storage is the main bottleneck.

\subsection{Accuracy and efficiency for high-order PDEs}
\label{sec:high-order-results}

\subsubsection{Benchmarks}

The high-order benchmarks comprise the 2D Korteweg--de Vries (KdV) equation, the 2D Kadomtsev--Petviashvili (KP) equation \cite{pu2024lax}, and a 1D gradient-enhanced KdV (g-KdV) equation \cite{yu22_gepinn}.
The required high-order and mixed derivatives use the Taylor-jet construction \cite{shi2024stochastic}; complete specifications are given in the supplementary material.

The three high-order PDE benchmarks are:
\begin{itemize}
  \item \textbf{2D Korteweg--de Vries (KdV):}
  \begin{equation}
    u_{ty} + u_{x x x y} + 3(u_{y}u_{x})_{x} - u_{x x} + 2u_{y y} = 0.
  \end{equation}
  \item \textbf{2D Kadomtsev--Petviashvili (KP):}
  \begin{equation}
    (u_{t} + 6 u u_{x} + u_{x x x})_{x} + 3\sigma ^{2} u_{y y} = 0,
  \end{equation}
  with expanded residual
  \begin{equation}
    u_{tx} + 6 u_{x} u_{x} + 6 u u_{xx} + u_{x x x x} + 3\sigma ^{2} u_{y y} = 0.
  \end{equation}
  \item \textbf{Gradient-enhanced 1D Korteweg--de Vries (g-KdV):}
  \begin{equation}
    u_{t} + u u_{x} + \alpha u_{x x x}=0.
  \end{equation}
  For gPINN, the residual and its gradient are
  \begin{equation}
  \begin{split}
    R(x,t) &:= u_{t} + u u_{x} + \alpha u_{x x x}, \\
    \nabla R(x,t) &= \mqty[u_{t t} + u_{t}u_{x} + uu_{t x} + \alpha u_{t x x x }, & u_{t x} + u_{x}u_{x} + uu_{xx} + \alpha u_{x x x x}].
  \end{split}
  \end{equation}
\end{itemize}


\subsubsection{Computational performance}

Tables~\ref{tab:2d-kdv}--\ref{tab:1d-gkdv} show a consistent accuracy--resource trade-off across the three high-order benchmarks.
Although the KdV, KP, and gradient-enhanced KdV problems involve different combinations of high-order and mixed derivatives, the relative computational behavior of SDZE remains similar across these cases.
Removing the reverse-mode parameter update reduces the parameter-side memory and computational overhead, while the resulting solutions remain close in accuracy to the strongest first-order references.
This consistency is important because spatial-derivative evaluation becomes substantially more involved for these equations.
The results therefore indicate that the matrix-free zeroth-order parameter update is not restricted to the second-order high-dimensional setting, but can also be coupled effectively with high-order spatial estimators.

The three benchmarks also expose the resulting accuracy--efficiency trade-off.
For the KdV-type problems, SDZE remains close to the most accurate first-order solutions while operating with a lighter parameter-update path.
The gradient-enhanced case provides a particularly informative comparison: the full-backpropagation reference retains an accuracy advantage, whereas SDZE favors lower computational cost while remaining close in solution quality.
These experiments do not imply that zeroth-order optimization universally dominates first-order training in approximation accuracy.
Rather, they show that a substantial fraction of first-order accuracy can be retained while moving the computation to a markedly different memory--runtime regime, which is particularly relevant for memory-constrained scientific computing.
\begin{table}[htbp]
  \footnotesize
\centering
\caption{Relative $L_2$ error, peak GPU memory, and iteration time for 2D KdV at $T=0.1$.}
\label{tab:2d-kdv}
{\renewcommand{\arraystretch}{1.15}
\resizebox{\textwidth}{!}{%
\begin{tabular}{lcccc}
\hline
Method & Relative $L_2$ Error $\downarrow$ & Memory (MB) $\downarrow$ & Time (s/it) $\downarrow$ & Backprop-Free \\ \hline
Backward Stacked (FO) & 3.21e-04 & 847 & 2.34 & No \\ \hline
Forward Jacobian (FO) &\textbf{2.98e-04} & 523 & 3.12 & No \\ \hline
STDE (FO) & 3.45e-04 & 312 & 1.87 & No \\ \hline
\textbf{SDZE (Ours, ZO)} & \underline{3.08e-04} & \textbf{287} & \textbf{1.52} & \textbf{Yes} \\ \hline
\end{tabular}}}
\end{table}

\begin{table}[htbp]
  \footnotesize
\centering
\caption{Relative $L_2$ error, peak GPU memory, and iteration time for 2D KP at $T=0.05$.}
\label{tab:2d-kp}
{\renewcommand{\arraystretch}{1.15}
\resizebox{\textwidth}{!}{%
\begin{tabular}{lcccc}
\hline
Method & Relative $L_2$ Error $\downarrow$ & Memory (MB) $\downarrow$ & Time (s/it) $\downarrow$ & Backprop-Free \\ \hline
Backward Stacked (FO) & 5.67e-04 & 1243 & 4.21 & No \\ \hline
Forward Jacobian (FO) & \textbf{5.23e-04} & 756 & 5.34 & No \\ \hline
STDE (FO) & 5.89e-04 & 445 & 3.12 & No \\ \hline
\textbf{SDZE (Ours, ZO)} & \underline{5.41e-04} & \textbf{398} & \textbf{2.67} & \textbf{Yes} \\ \hline
\end{tabular}}}
\end{table}

\begin{table}[!ht]
  \footnotesize
\centering
\caption{Relative $L_2$ error, peak GPU memory, and iteration time for g-KdV with gradient regularization.}
\label{tab:1d-gkdv}
{\renewcommand{\arraystretch}{1.15}
\resizebox{\textwidth}{!}{%
\begin{tabular}{lcccc}
\hline
Method & Relative $L_2$ Error $\downarrow$ & Memory (MB) $\downarrow$ & Time (s/it) $\downarrow$ & Backprop-Free \\ \hline
Backward Stacked (FO) & 2.14e-04 & 234 & 0.87 & No \\ \hline
Forward Jacobian (FO) & 1.98e-04 & 198 & 1.12 & No \\ \hline
STDE (FO) & 2.31e-04 & 156 & 0.72 & No \\ \hline
gPINN-BP (FO) & \textbf{1.67e-04} & 412 & 1.45 & No \\ \hline
\textbf{SDZE-gPINN (Ours)} & \underline{1.82e-04} & \textbf{143} & \textbf{0.68} & \textbf{Yes} \\ \hline
\end{tabular}}}
\end{table}

\begin{figure}[!htb]
  \centering
  \includegraphics[width=.8\textwidth]{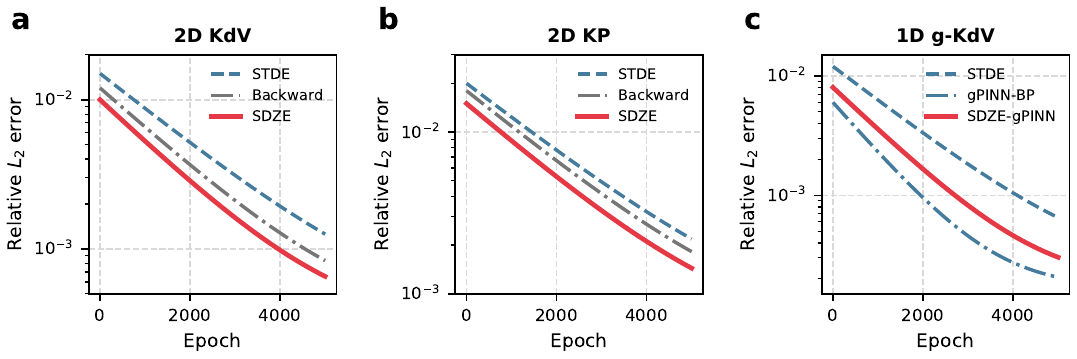}
  \vspace{-8pt}
  \caption{Relative $L_2$ error trajectories during training for 2D KdV, 2D KP, and 1D g-KdV.}
  \label{fig:high-order-convergence}
\end{figure}

The training trajectories in Figure~\ref{fig:high-order-convergence} provide further evidence that the table-level observations are not isolated endpoint effects.
SDZE maintains a competitive decrease of the solution error on all three high-order problems, while the relative ordering varies across the PDEs.
This variation indicates that SDZE's advantage lies in the computational structure of the parameter update rather than universal superiority in optimization accuracy.

\begin{figure}[!htb]
  \centering
  \includegraphics[width=\textwidth]{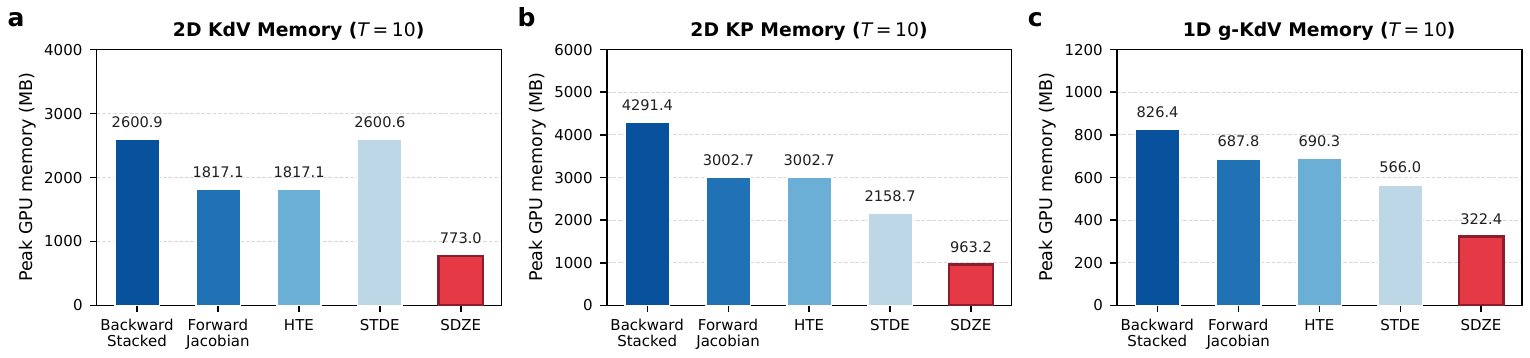}
  \vspace{-16pt}
  \caption{Peak GPU memory across methods for 2D KdV, 2D KP, and 1D g-KdV with $T=10$.}
  \label{fig:high-order-memory}
\end{figure}

Figure~\ref{fig:high-order-memory} further shows that the memory ordering is preserved when the temporal workload is enlarged.
Because this behavior appears across distinct high-order operators, it is consistent with a parameter-side mechanism rather than with a particular PDE or short-horizon configuration.
Taken together, the quantitative comparisons in Tables~\ref{tab:2d-kdv}--\ref{tab:1d-gkdv} and the trajectory and long-horizon results in Figures~\ref{fig:high-order-convergence}--\ref{fig:high-order-memory} show that the accuracy--memory--runtime trade-off observed in the high-dimensional experiments extends to high-order and mixed-derivative PDEs.

\subsection{Bias--variance behavior and training stability}

\begin{figure}[htbp]
  \centering
  \includegraphics[width=.93\textwidth]{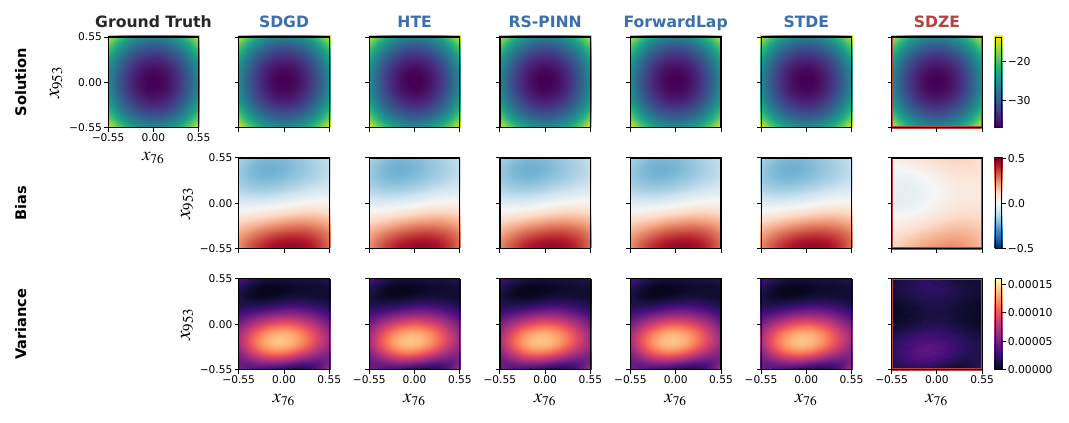}
  \vspace{-13pt}
  \caption{Solution, bias, and spatial variance on the 1000-D Allen--Cahn slice $(x_{76},x_{953})$.}
  \label{fig:1000d-bias-spatial-variance-cross-scale}
\end{figure}

\begin{figure}[htbp]
  \centering
  \includegraphics[width=.93\textwidth]{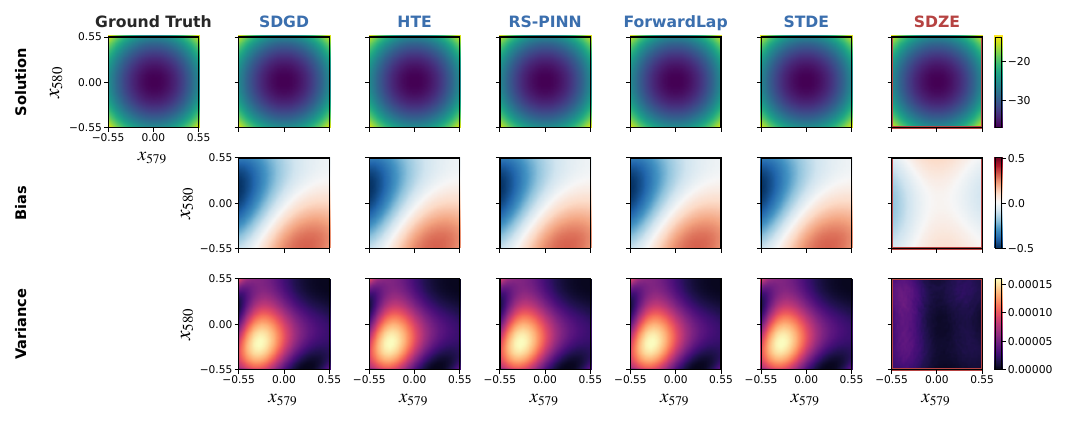}
  \vspace{-13pt}
  \caption{Solution, bias, and spatial variance on the 1000-D Allen--Cahn slice $(x_{579},x_{580})$.}
  \label{fig:1000d-bias-spatial-variance-medium-near}
\end{figure}

Figure~\ref{fig:1000d-bias-spatial-variance-cross-scale} compares prediction bias and seed variance on the widely separated slice $(x_{76},x_{953})$, whereas Figure~\ref{fig:1000d-bias-spatial-variance-medium-near} reports the adjacent slice $(x_{579},x_{580})$.
Across both slice pairs, SDZE shows weaker signed bias and lower spatial variance than the stochastic first-order baselines; these solution-level diagnostics are consistent with CRNS stabilization but distinct from the conditional estimator variance in Proposition~\ref{prop:crns}.

\vspace{-12pt}

\FloatBarrier

\subsection{Ablation studies}\label{sec:zo_ablations}

The ablations isolate CRNS, subspace rank, and refresh frequency on the 10K-D Allen--Cahn benchmark.

\subsubsection{CRNS and optimization stability}

Independent spatial states in the two directional evaluations introduce the $\mathcal O(\epsilon^{-2})$ term identified in Proposition~\ref{prop:crns}.
To isolate CRNS, the network parameters, direction, perturbation radius, collocation points, and spatial-oracle budget are fixed, while only the spatial state is resampled.
Table~\ref{tab:crns_ablation} shows that independent states produce variance of order $10^8$ and unstable training, whereas CRNS keeps the variance near $10^{-2}$ and yields stable runs.

\begin{table}[htbp]
\centering
\caption{Effect of CRNS on optimization stability.}
\label{tab:crns_ablation}
\resizebox{0.45\textwidth}{!}{
\begin{tabular}{lcc}
\hline
Method & $\widehat{\operatorname{Var}}_s(\hat\delta)$ & Status \\ \hline
SDZE (w/o CRNS) & $\sim 10^{8}$ (Large) & Unstable \\
\textbf{SDZE (w/ CRNS)} & $\mathbf{\sim 10^{-2}}$ (Bounded) & \textbf{Stable} \\ \hline
\end{tabular}}
\end{table}


\subsubsection{Comparison with zeroth-order methods}

\begin{figure}[!htb]
  \centering
  \includegraphics[width=.95\textwidth]{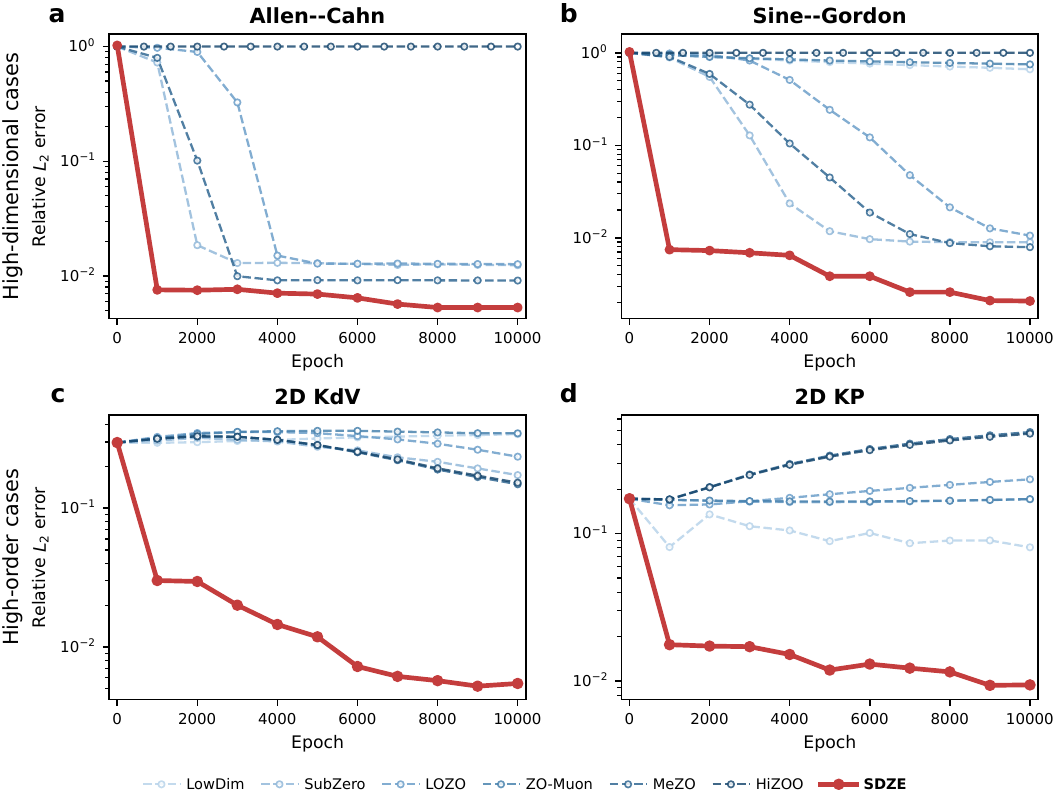}
  \vspace{-8pt}
  \caption{Relative $L_2$ error during training for SDZE and zeroth-order baselines.}
  \label{fig:llm-zo-vs-sdze}
\end{figure}

Figure~\ref{fig:llm-zo-vs-sdze} compares SDZE with LowDim, SubZero, LOZO, ZO-Muon, MeZO, and HiZOO~\cite{malladi2023fine,yu2025zeroth,zhang2024revisiting} on two high-dimensional and two high-order problems.
SDZE continues to reduce the error throughout training, whereas the compared zeroth-order baselines plateau at larger errors or become unstable.
Because the methods differ in both random-state coupling and perturbation geometry, this comparison is not an isolated CRNS ablation.

\subsubsection{Variance scaling under CRNS}

\begin{figure}[htbp]
  \centering
  \includegraphics[width=\textwidth]{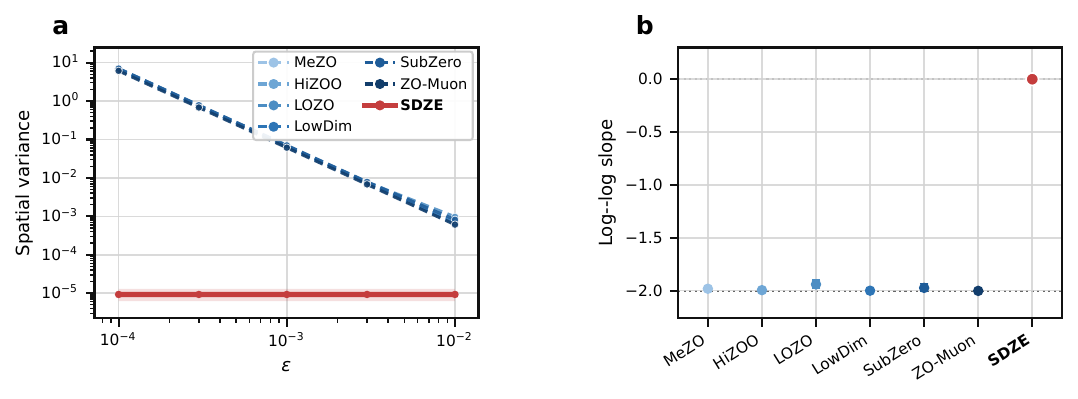}
  \vspace{-18pt}
  \caption{Conditional spatial variance of independent zeroth-order controls and SDZE with CRNS.}
  \label{fig:sdze_crns_independent_zo_variance}
\end{figure}

To isolate the spatial component of the directional estimator, this test fixes the network, parameter direction, collocation points, and spatial-oracle budget for Allen--Cahn.
The controls use independent spatial states for the two residual evaluations, whereas SDZE reuses the same state through CRNS.
In Figure~\ref{fig:sdze_crns_independent_zo_variance}, \textbf{a} shows that the conditional spatial variance of the independent controls increases as $\epsilon$ decreases, following the inverse-square trend.
In contrast, the CRNS variance remains near $10^{-5}$ and shows no apparent dependence on $\epsilon$.
The log--log slopes in \textbf{b} confirm $\mathcal O(\epsilon^{-2})$ scaling for the independent controls and a near-zero slope for SDZE.
This evidence is consistent with Proposition~\ref{prop:crns} and the stability contrast in Table~\ref{tab:crns_ablation}.

\FloatBarrier



\section{Conclusions} 
\label{sec:conclusion}
In this work, the SDZE is introduced for PINN solvers for high-dimensional and high-order PDEs.
Unlike conventional zeroth-order updates, SDZE applies parameter perturbations to stochastic spatial residual evaluations.
Under the stated regularity conditions, CRNS removes the independent-sampling $\mathcal{O}(\varepsilon^{-2})$ variance term, while the experiments show stable training in the tested settings.
The matrix-free update avoids an explicit global subspace basis, dense parameter-perturbation matrices, and reverse-mode parameter-gradient storage, while retaining the unavoidable $\mathcal{O}(P)$ storage of the parameters.
The theoretical analysis separates spatial randomness, Gaussian direction randomness, and incomplete subspace coverage.
The analysis characterizes the residual-gradient noise after CRNS and establishes biases and variance bounds for the projected pseudo-gradients.
The finite-time projected-stationarity bound does not require full-space coverage, whereas the full-gradient bounds require explicit coverage and refresh conditions.
Experiments show stable training and solution accuracy comparable to executable stochastic first-order baselines, together with lower peak GPU memory and a wider executable dimension range.
The ablation results further confirm the role of CRNS and reveal the trade-off between the subspace rank and the refresh frequency.
These results support SDZE as a memory-efficient option when reverse-mode AD exceeds the available device memory.


\section*{Data and code availability}
The source code and the data-generation and evaluation scripts are publicly available at \url{https://github.com/liangzhangyong/SDZE}.


\clearpage
\appendix
\mathtoolsset{showonlyrefs=false}

\section{Extended related work}
\label{sec:supp-related}

\subsection{High-order automatic differentiation}
Forward-mode AD has long supported high-order spatial derivatives \cite{bendtsen1997tad,karczmarczuk1998ad}.
Later works develop more general abstractions for higher-order differentiation \cite{wang2017higherad,laurel2022higherad}.
Taylor-mode AD is now accessible in machine-learning workflows through JAX-based implementations \cite{bettencourt2019taylor,jax2018github}.
Operator-specific forward rules have also been developed for the Laplacian \cite{li2024computational} and related differential operators \cite{li24_dof}.
Other work randomizes the linearized portion of an AD computation graph \cite{oktay2021randomad}, while forward-mode AD can also be used to compute the parameter gradients \cite{baydin22_gradien_backp}.
Coupled automatic--numerical differentiation has likewise been utilized to accelerate PINN residual evaluation while retaining derivative information \cite{chiu2022can}.
Smoothing-kernel gradients provide another approximation to spatial derivatives without relying on conventional coordinate-wise AD \cite{pan2025sk}.

\subsection{Randomized spatial-operator estimation}
To reduce the cost of high-dimensional spatial operators, randomized methods are widely used in numerical linear algebra \cite{martinsson2020randomized,murray2023rnla}.
Random projections provide one route to dimension reduction through the Johnson--Lindenstrauss principle \cite{ghojogh2021jl}.
Amortized optimization instead learns mappings from problem instances to solutions \cite{amos2023tutorial}.
For PINNs, SDGD reduces the memory cost of additive spatial operators by sampling a subset of their terms at each iteration \cite{hu2024sdgd}.
Stochastic spatial estimators have also been developed for fractional and tempered fractional operators, in which Monte Carlo sampling replaces direct evaluations of high-dimensional integral derivatives \cite{hu2024tackling,guo2022monte}.
Hutchinson's trace estimator provides another approach to randomized operator estimation \cite{hutchinson1989trace}, which has been used in diffusion models \cite{song2019slicedscore} and in PINNs \cite{hu2024hutch}.
Variance-reduced stochastic first-order methods provide a complementary analysis of how the mean accuracy of a stochastic gradient estimator affects the training convergence \cite{driggs2022accelerating}.
Beyond stochastic operator estimators, randomized-feature methods fix a randomized representation and solve for a reduced set of coefficients.
High-dimensional neural surrogates such as Anant-Net target scalable and interpretable PDE approximations without forming the full parameter-gradient graph \cite{menon2025anant}, while separable PI-DeepONet reduces the effective dimensionality of PINNs through separable operator representations \cite{mandl2025separable}.

\subsection{Zeroth-order parameter optimization}
At the parameter-update level, classical stochastic optimization treats both first-order and zeroth-order access models for nonconvex objectives \cite{ghadimi2013stochastic}.
ZO methods use function evaluations when derivatives are unavailable or costly \cite{liu2020primer}, including memory-constrained large-language-model training \cite{malladi2023fine}.
Complementary PINN optimization strategies modify the loss geometry directly by minimizing curvature and flattening the residual landscape \cite{moayedi2025cupinn}.
These methods retain first-order parameter optimization rather than using a stochastic zeroth-order update.
Orthogonal random directions and approximately sparse gradients provide two routes for reducing the ambient-dimension dependence of ZO estimation \cite{kozak2023zeroth,cai2022zeroth}.
Gradient-free PINN training has also been studied in the stochastic-projection SP-PINN framework \cite{navaneeth2023stochastic}.
Unlike backpropagation-based SGD \citep{amari1993backpropagation} and Adam \citep{kingma15_adam}, two-point ZO updates do not require a reverse-mode parameter-gradient graph.
Convergence guarantees cover several ZO access models \citep{nesterov2017random,duchi2015optimal,liu2018zeroth,ji2019improved}.
Derivative-free random-subspace methods distinguish active from ambient parameter dimensions \citep{cartis2023scalable,nozawa2024zeroth,roberts2023direct,kozak2021stochastic}.
Existing large-model approaches also bring in larger batches \citep{gautamvariance,jiang2024zo}, sparse perturbations \citep{liu2026sparse,zhang2024revisiting}, and parameter-efficient architectures \citep{yang2024adazeta}.
Sample-path methods reuse common random numbers in paired stochastic evaluations \cite{robinson1996analysis}.
Other methods draw on unbiased multilevel Monte Carlo gradients and low-rank randomized partial-trace estimators \cite{goda2022unbiased,chen2024faster}.
SDZE combines matrix-free layer-wise perturbations, which avoid an explicit global subspace basis, with complete-state CRNS, which removes the independent-state $\mathcal O(\epsilon^{-2})$ variance amplification in paired stochastic residual evaluations.

\section{Estimator preliminaries}
\label{sec:supp-random-gradient}

\subsection{First-order automatic differentiation}
\label{sec:supp-first-order-gradient-estimation}
First-order automatic differentiation (AD) computes gradients by differentiating compositions of elementary functions.
We present the neural network $F_{\boldsymbol{\theta}}: \mathbb{R}^{d} \to \mathbb{R}^{d'}$ as a composition of primitive functions $F_i$, each parameterized by $\boldsymbol{\theta}_i$.
Consider a linear computation graph of the form $F = F_L \circ F_{L-1} \circ \dots \circ F_1$ with uniform hidden dimension $h$.
For this computation graph, derivatives can be evaluated by AD in forward or backward mode.

In forward-mode AD, each primitive $F_i$ is linearized by the Fr\'echet derivative $\partial F_i: \mathbb{R}^h \to \mathrm{L}(\mathbb{R}^h, \mathbb{R}^h)$.
At a primal point $\boldsymbol{a}$ and in a tangent direction $\boldsymbol{v}$, the derivative gives the Jacobian-vector product (JVP) $\partial F_i(\boldsymbol{a})(\boldsymbol{v}) = \left. \frac{\partial F_i}{\partial \boldsymbol{x}} \right|_{\boldsymbol{a}} \boldsymbol{v}$.
Composing the linearized primitives gives
\begin{equation}
\frac{\partial F}{\partial \boldsymbol{x}} \boldsymbol{v} = [\partial F_L \circ \partial F_{L-1} \circ \dots \circ \partial F_1](\boldsymbol{x})(\boldsymbol{v}).
\end{equation}

Computing the full Jacobian requires $d$ JVPs; sequential JVP evaluation uses $\mathcal{O}(\max(d,h))$ working memory, excluding stored parameters and the assembled Jacobian.
By contrast, backward-mode linearization is defined by the adjoint of the Fr\'echet derivative, $\partial^\top F_i$.
At $\boldsymbol{a}$, the vector-Jacobian product (VJP) maps a cotangent $\boldsymbol{v}^\top$ to
\[
  \partial^\top F_i(\boldsymbol{a})(\boldsymbol{v}^\top)
  =\boldsymbol{v}^\top
    \left.\frac{\partial F_i}{\partial\boldsymbol{x}}\right|_{\boldsymbol{a}}.
\]
The reverse-order backward-mode calculation requires a preceding forward pass that caches the evaluation trace $\{\boldsymbol{y}_i\}_{i=1}^L$.
The corresponding memory requirement is $\mathcal{O}(d + (L-1)h)$.
Although efficient for scalar cost functions, nested first-order AD for high-order input derivatives can substantially increase memory and computation; the cost depends on the derivative representation and AD implementation.

\subsection{Zeroth-order parameter estimation}
\label{sec:supp-zo-gradient}

Two-point ZO estimation replaces reverse-mode parameter differentiation with paired function evaluations.
For a minibatch $\mathcal{B}$, we use the following two-point Gaussian ZO estimator:
\begin{equation}
\widehat{\nabla} \mathcal{L}(\boldsymbol{\theta}; \mathcal{B}) = \frac{\mathcal{L}(\boldsymbol{\theta} + \epsilon \boldsymbol{z}; \mathcal{B}) - \mathcal{L}(\boldsymbol{\theta} - \epsilon \boldsymbol{z}; \mathcal{B})}{2 \epsilon} \boldsymbol{z},
\label{eq:sm-mezo-estimate-grad}
\end{equation}
where $\boldsymbol{z} \in \mathbb{R}^P \sim \mathcal{N}(\boldsymbol{0}, \boldsymbol{I}_P)$ is the perturbation vector and $\epsilon > 0$ is the perturbation scale.
The estimator in \eqref{eq:sm-mezo-estimate-grad} is unbiased for the gradient of the Gaussian-smoothed objective.
Each update requires two forward evaluations and has the form $\boldsymbol{\theta}^{t+1}=\boldsymbol{\theta}^{t}-\alpha_t\widehat{\nabla}\mathcal{L}(\boldsymbol{\theta}^{t};\mathcal{B}^t)$.

However, isotropic perturbations over the full parameter space $\mathbb{R}^P$ have gradient variance that scales as $\mathcal{O}(P)$.
This can destabilize PINN training for stiff PDEs when the network contains massive weight matrices.
Restricting perturbations to low-dimensional random subspaces is able to reduce the variance contribution \citep{nozawa2024zeroth, roberts2023direct}.
Let $\mathcal{Q} \in \mathbb{R}^{P \times q}$ be an orthonormal basis with $q \ll P$.
The perturbations are projected as follows:
\begin{equation}
\tilde{\boldsymbol{z}} = \mathcal{Q} \boldsymbol{z}, \quad \boldsymbol{z} \in \mathbb{R}^q \sim \mathcal{N}(\boldsymbol{0}, \boldsymbol{I}_q),
\end{equation}

This gives the subspace two-point Gaussian ZO estimator:
\begin{equation}
\widehat{\nabla} \mathcal{L}(\boldsymbol{\theta}, \mathcal{Q}; \mathcal{B}) = \frac{\mathcal{L}(\boldsymbol{\theta} + \epsilon \mathcal{Q} \boldsymbol{z}; \mathcal{B}) - \mathcal{L}(\boldsymbol{\theta} - \epsilon \mathcal{Q} \boldsymbol{z}; \mathcal{B})}{2 \epsilon} \mathcal{Q} \boldsymbol{z}.
\label{eq:sm-subspace-zo-estimate-grad}
\end{equation}

For the estimator in \eqref{eq:sm-subspace-zo-estimate-grad}, the direction-sampling contribution to the variance bound depends on the subspace dimension $q$, rather than on the full parameter dimension $P$.
However, storing an explicit basis $\mathcal{Q} \in \mathbb{R}^{P \times q}$ may be impractical for PINNs with $P \sim 10^7$ parameters.

\subsection{Stochastic dimension sampling}
\label{sec:supp-sdgd-review}
SDGD \cite{hu2024sdgd} samples terms of high-dimensional additive differential operators.
For $\mathcal{D} = \sum_{j=1}^{N_{\mathcal{D}}} \mathcal{D}_j$, the SDGD estimator is
\begin{equation}
\tilde{\mathcal{D}}_J = \frac{N_{\mathcal{D}}}{|J|} \sum_{j \in J} \mathcal{D}_j,
\end{equation}
where $J$ is sampled uniformly and $|J|$ is the number of sampled operator terms.
During the automatic differentiation, non-sampled input dimensions are treated as constants.
The amortized memory complexity is $\mathcal{O}(|J| \cdot 2^{k-1}(1 + (L-1)h))$, rather than $\mathcal{O}(2^{k-1}(d + (L-1)h))$.
The linear memory dependence is on $|J|$ rather than on $d$.
The exponential dependence on the derivative order $k$ remains.

\def\SDZEProofs{}

\section{Computational and experimental details}
\subsection{Implementation and evaluation protocol}

Unless otherwise noted, runnable implementations use JAX on a single NVIDIA GeForce RTX 5090 GPU.
The extreme-dimensional Allen--Cahn memory study uses two NVIDIA GeForce RTX 5090 GPUs with 64~GB total memory.
For each benchmark, we use a fully connected PINN with 4 layers, 128 hidden units per hidden layer, and $\tanh$ activations.
The first-order baselines use Adam, whereas SDZE applies the direct pseudo-gradient update in \eqref{eq:delta} without parameter-wise moment states.
For the resource-profiling experiments, the initial step size is $\alpha_0=1\times10^{-3}$ and follows the linear schedule
\[
\alpha_t=\alpha_0\left(1-\frac{t}{N_{\mathrm{prof}}}\right),
\qquad N_{\mathrm{prof}}=256,
\]
for $t=0,\ldots,N_{\mathrm{prof}}-1$.
Each iteration uses 100 residual points and up to 16 sampled spatial probes.
This probe count applies to each internal copy in the cross-sampled loss.
One SDZE parameter direction uses two loss queries, and the complete pair of internal probe states is reused at the positive and negative queries.
The STDE/SDZE spatial batch size is $\min(d,16)$.
SDGD and FOBAD use 4 sampled directions, HTE uses 16 directions, RS-PINN uses 128 samples, and ForwardLap computes the exact Laplacian.
The 256-iteration budget applies to resource profiling; the longer convergence experiments use the iteration ranges shown on the corresponding axes.
OOM denotes a run that exceeds the available memory of the corresponding hardware configuration.
Within each benchmark, the architecture, spatial batch sizes, random seeds, iteration count, and evaluation grid are held fixed across compared methods unless a method-specific requirement is stated.
Accuracy measurements use $S=5$ independent seeds, and the main tables report the sample mean.
Within each process-level run, iteration times are averaged over 32 post-warm-up iterations.
Peak memory is recorded as the maximum allocated GPU memory for single-GPU runs and the peak aggregate allocation for the two-GPU extreme-memory study.
For interpretation, the peak allocation is decomposed as
\begin{equation}
M_{\mathrm{peak}}
=\max_{\tau}\left\{
M_{\mathrm{model}}+M_{\mathrm{optimizer}}
+M_{\mathrm{grad/VJP}}(\tau)+M_{\mathrm{spatial}}(\tau)
+M_{\mathrm{update}}(\tau)
\right\},
\label{eq:peak-memory-components}
\end{equation}
In \eqref{eq:peak-memory-components}, $\tau$ indexes operations within a training iteration.
Thus, the reported total-memory difference between the first-order baselines and SDZE includes both the removal of reverse-mode gradient/VJP storage and the absence of Adam moment states; these contributions are not isolated without component-level allocation traces.
The reported timing is the median over 3 independent process-level runs.
For the relative-error metric, seed-level values $E_{L_2}^{(s)}$ have sample mean and standard deviation
\begin{equation}
\overline E_{L_2}=\frac1S\sum_{s=1}^{S}E_{L_2}^{(s)},
\qquad
s_E=\left[
\frac1{S-1}\sum_{s=1}^{S}
\left(E_{L_2}^{(s)}-\overline E_{L_2}\right)^2
\right]^{1/2}.
\end{equation}

\subsection{Rank and refresh sensitivity}

On the 10K-D Allen--Cahn benchmark, we vary the nominal rank $r$ and refresh frequency $F$, with the feasible layer-wise ranks set to $r_l=\min\{r,m_l,n_l\}$.
Table~\ref{tab:abla-rank-T} reports the resulting relative $L_2$ errors.
Among the settings in Table~\ref{tab:abla-rank-T}, $r=128$ and $F=500$ give the lowest error on this benchmark.
This comparison is restricted to the settings and benchmark reported in Table~\ref{tab:abla-rank-T}.
Larger ranks increase directional coverage, whereas infrequent refreshes reduce subspace exploration and lead to larger optimization errors.

\begin{table}[htbp]
\centering
\caption{Relative $L_2$ error for 10K-D Allen--Cahn versus refresh frequency $F$ and rank $r$.}
\label{tab:abla-rank-T}
\resizebox{0.45\textwidth}{!}{
\begin{tabular}{lccc}
\hline
$F$ \textbackslash~$r$ & 32 & 64 & 128 \\ \hline
$500$ & 1.23e-2 & 8.56e-3  & \textbf{4.21e-3}\\
$1000$& 2.06e-2 & 1.45e-2 & 9.87e-3\\
$2000$ & 3.45e-2 & 2.12e-2 & 1.56e-2\\ \hline
\end{tabular}}
\end{table}

\subsection{PINN and stochastic-residual formulation}

PINNs \cite{raissi2019pinn} approximate the solution by a neural-network output \(u_{\theta }(\vb{x})\).
For the spatial PDE benchmarks, we consider problems on \(\Omega\subset\mathbb{R}^d\) of the form
\begin{equation}\label{eqn:pde}
\begin{aligned}
\mathcal{N} u(\vb{x}) &= f(\vb{x}), \quad \vb{x} \in \Omega, \\
\mathcal{D} u(\vb{x}) &= g(\vb{x}), \quad \vb{x} \in \partial\Omega.
\end{aligned}
\end{equation}
where \(\mathcal{N}\) and \(\mathcal{D}\) denote the PDE and boundary operators, respectively, and $f(\vb{x})$ and $g(\vb{x})$ specify the source and boundary or initial data.
We approximate the unknown solution \(u:\mathbb{R}^d\to\mathbb{R}\) by minimizing the mean-squared residual
\begin{equation} \label{eqn:pinn-loss}
  \ell _{\text{residual}}(\theta; \{\vb{x}^{(i)}\}_{i=1}^{N_{r}})
  =   \frac{1}{N_{r}} \sum_{i=1}^{N_{r}} \abs{\mathcal{N}u_{\theta}(\vb{x}^{(i)}) - f(\vb{x}^{(i)})}^{2}.
\end{equation}
The collocation points \(\{\vb{x}^{(i)}\}_{i=1}^{N_{r}}\) are sampled in \(\Omega\).
Following \cite{lu2021hardpinn}, we reparameterize $u_\theta$ so that \(\mathcal{D}u_\theta(\vb{x})=g(\vb{x})\) on \(\partial\Omega\); the resulting objective therefore contains only the PDE residuals.

For additively decomposable differential operators, stochastic spatial estimators reduce the cost of evaluating the PDE residual.
For the Allen--Cahn equation, $\mathcal{N}u=\laplacian u+u-u^{3}$; the Laplacian is randomized while the nonlinear reaction term is evaluated exactly.
The corresponding cross-sampled objective estimator is
\begin{equation} \label{eqn:pinn-loss-stde}
  \tilde{\ell }_{\text{residual}}(\theta; \{\vb{x}^{(i)}\}_{i=1}^{N_{r}}, I, J) = \frac{1}{N_{r}} \sum_{i=1}^{N_{r}} \left[\tilde{\mathcal{N}} _{I} u_{\theta}(\vb{x}^{(i)}) - f(\vb{x}^{(i)})\right] \cdot \left[\tilde{\mathcal{N}} _{J} u_{\theta}(\vb{x}^{(i)}) - f(\vb{x}^{(i)})\right].
\end{equation}
The estimator \eqref{eqn:pinn-loss-stde} replaces the squared residual in \eqref{eqn:pinn-loss} with the product of two independently drawn spatial-operator estimators.
Under the domination condition \eqref{eq:det-dominated-gradient}, differentiation and expectation can be interchanged, so the expected gradient equals the gradient of the residual objective.
SDZE estimates parameter gradients from two forward residual evaluations in a low-dimensional subspace and therefore avoids reverse-mode parameter differentiation.

\subsection{High-dimensional PDE benchmarks}

For auxiliary benchmarks, the following manufactured solutions use $c_i\sim\mathcal N(0,1)$:
\begin{itemize}
  \item two-body: $u_{\mathrm{exact}}(\vb{x})=(1-\|\vb{x}\|_2^2)\sum_{i=1}^{d-1}c_i\exp(x_i x_{i+1})$;
  \item three-body: $u_{\mathrm{exact}}(\vb{x})=(1-\|\vb{x}\|_2^2)\sum_{i=1}^{d-2}c_i\exp(x_i x_{i+1}x_{i+2})$.
\end{itemize}

\subsection{Adopted STDE Taylor-jet specifications}

The following Taylor-jet constructions follow STDE~\cite{shi2024stochastic} and specify only the spatial-derivative evaluation used in the high-order benchmarks.
They are not part of the SDZE parameter-update mechanism.
The jet order is the order of the Taylor-mode computational encoding rather than the differential order of the underlying PDE.
We use derivative-valued jets: for an input jet $(\vb{v}_1,\ldots,\vb{v}_K)$,
\[
\mathfrak J_{[k]}=\left.\frac{d^k}{ds^k}
u\!\left(\vb{x}+\sum_{j=1}^{K}\vb{v}_j\frac{s^j}{j!}\right)\right|_{s=0}.
\]
Thus $[k]$ denotes derivative order $k$, corresponding to index $k-1$ in a zero-based array of derivative outputs.
The identities below are stated under this convention.

\subsubsection{2D KdV}

For the 2D KdV benchmark, STDE uses the Taylor-jet push-forwards
\begin{equation}
  \mathfrak{J}^{(1)}=\dd^9 u(\vb{x}, \vb{0}, \vb{e}_{x}, \vb{e}_{y}, \vb{0}, \dots), \;\;
\mathfrak{J}^{(2)}=\dd^3 u(\vb{x}, \vb{e}_{y}, \vb{e}_{t}, \vb{0}), \;\;
\mathfrak{J}^{(3)}=\dd^3 u(\vb{x}, \vb{e}_{y}, \vb{0}, \vb{0}).
\end{equation}
The required derivative identities are
\begin{equation}
\begin{aligned}
  u_{x}  = \mathfrak{J}^{(1)}_{[2]}, \;
  u_{y}  = \mathfrak{J}^{(1)}_{[3]}, \;
  u_{x x}  = \mathfrak{J}^{(1)}_{[4]}  / 3 , \;
  u_{x y}  = \mathfrak{J}^{(1)}_{[5]}  / 10 ,
  u_{y y}  = \mathfrak{J}^{(3)}_{[2]}  ,  \\
  u_{y y y}  = \mathfrak{J}^{(3)}_{[3]}  ,
  u_{x x x y} = (\mathfrak{J}^{(1)}_{[9]} - 280 u_{y y y} ) / 1260, \;
  u_{t y} = (\mathfrak{J}^{(2)}_{[3]} -  u_{y y y} ) / 3.
\end{aligned}
\end{equation}

\subsubsection{2D KP}

For the 2D KP benchmark, STDE uses the 5-jet, 4-jet, and 2-jet push-forwards
\begin{equation}
\mathfrak{J}^{(1)}:=\dd^5 u(\vb{x}, \vb{0}, \vb{e}_{t}, \vb{e}_{x}, \vb{0}, \vb{0}),\quad
\mathfrak{J}^{(2)}:=\dd^4 u(\vb{x}, \vb{e}_{x}, \vb{0}, \vb{0}, \vb{0}),\quad
\mathfrak{J}^{(3)}:=\dd^2 u(\vb{x}, \vb{e}_{y}, \vb{0}).
\end{equation}
The associated derivative identities are
\begin{equation}
\begin{aligned}
  u_{t x}  = \mathfrak{J}^{(1)}_{[5]}  / 10 ,  \\
  u_{x}  = \mathfrak{J}^{(2)}_{[1]}, \;
  u_{x x}  = \mathfrak{J}^{(2)}_{[2]} , \;
  u_{x x x x}  = \mathfrak{J}^{(2)}_{[4]} ,  \\
  u_{y y} = \mathfrak{J}^{(3)}_{[2]}.
\end{aligned}
\end{equation}

\subsubsection{1D g-KdV}

For the 1D g-KdV benchmark, STDE uses the 2-jet and 7-jet push-forwards
\begin{equation}
\resizebox{\linewidth}{!}{$\displaystyle
\mathfrak{J}^{(1)}:=\dd^{7} u(\vb{x}, \vb{e}_{x}, \vb{0}, \vb{0}, \vb{0}, \vb{0}, \vb{0}, \vb{0}),\quad
\mathfrak{J}^{(2)}:=\dd^7 u(\vb{x}, \vb{e}_{x}, \vb{0}, \vb{0}, \vb{e}_{t}, \vb{0}, \vb{0}, \vb{0}),\quad
\mathfrak{J}^{(3)}:=\dd^2 u(\vb{x}, \vb{e}_{t}, \vb{0}).
$}
\end{equation}
The corresponding derivative identities are
\begin{equation}
\begin{aligned}
  u_{x}  = \mathfrak{J}^{(1)}_{[1]}, \;
  u_{x x}  = \mathfrak{J}^{(1)}_{[2]}, \;
  u_{x x x}  = \mathfrak{J}^{(1)}_{[3]}, \;
  u_{x x x x}  = \mathfrak{J}^{(1)}_{[4]}, \;
  u_{x x x x x}  = \mathfrak{J}^{(1)}_{[5]}, \\
  u_{t x x x}  = (\mathfrak{J}^{(2)}_{[7]} - \mathfrak{J}^{(1)}_{[7]}) / 35, \;
  u_{tx}  = (\mathfrak{J}^{(2)}_{[5]} - u_{x x x x x}) / 5, \;
  u_{t}  = \mathfrak{J}^{(2)}_{[4]} - u_{x x x x},  \\
  u_{t t} = \mathfrak{J}^{(3)}_{[2]}.
\end{aligned}
\end{equation}
The auxiliary pure derivative $u_{x x x x x}$ appears only in the coefficient elimination for this Taylor-jet encoding; neither the g-KdV equation nor its gradient-enhanced residual contains a fifth-order differential term.

\subsection{gPINN loss for the 1D g-KdV benchmark}

The gPINN \cite{yu22_gepinn} adds a residual-gradient penalty in this benchmark.
The residual gradient loss is
\begin{equation}
  \ell_{\text{gPINN}}
  =\frac{1}{N_r}\sum_{i=1}^{N_r}
  \left\|\nabla_{(t,x)}R\!\left(x^{(i)},t^{(i)}\right)\right\|_2^2.
\end{equation}
The total loss is
\begin{equation}
  \ell_{\text{total}} = \ell_{\text{residual }} + c_{\text{gPINN }}  \ell_{\text{gPINN }},
\end{equation}
where $c_{\text{gPINN }}$ is the gPINN penalty weight.

\subsection{Solution-level bias and seed-variance diagnostics}
\label{sec:supp-bias-variance-stability}

For predictions $u^{(s)}(\vb{x})$ from $S=5$ independent seeds, the solution-level mean, bias, and seed variance are
\begin{equation}
\begin{aligned}
\overline u(\vb{x})
&=\frac1S\sum_{s=1}^{S}u^{(s)}(\vb{x}),\\
\operatorname{Bias}(\vb{x})
&=\overline u(\vb{x})-u_{\mathrm{ref}}(\vb{x}),\\
\widehat{\operatorname{Var}}(\vb{x})
&=\frac1{S-1}\sum_{s=1}^{S}
\left(u^{(s)}(\vb{x})-\overline u(\vb{x})\right)^2.
\end{aligned}
\end{equation}

These are solution-level statistics across independently trained models, not the conditional estimator variance studied in Proposition~\ref{prop:crns} of the main manuscript.
In particular, lower seed variance does not by itself verify the CRNS variance bound at fixed parameters and perturbation direction.

\bibliographystyle{siamplain} \bibliography{references}
\end{document}